\documentclass[runningheads]{llncs}

\usepackage{accv}

\usepackage{accvabbrv}
\usepackage{graphicx}
\usepackage{booktabs}
\usepackage{colortbl}
\usepackage{soul}
\usepackage{multirow}
\usepackage{bbm}
\newcommand{\indicator}{\mathbbm{1}}
\usepackage[accsupp]{axessibility}  
\usepackage{algorithm}
\usepackage{algpseudocode}
\algrenewcommand{\algorithmicrequire}{\textbf{Input:}}
\algrenewcommand{\algorithmicensure}{\textbf{Output:}}
\algnewcommand{\Parameters}{\item[\textbf{Parameters:}]}

\usepackage{hyperref}

\usepackage{orcidlink}
\usepackage{comment}

\begin{document}

\title{Beyond Waypoint Regression: Query-Based Cost Learning over Reachable Ego Futures \\ for End-to-End Driving} 

\titlerunning{Beyond Waypoint Regression: Query-Based Cost for End-to-End Driving}

\author{Ahmed Abouelazm\inst{1,2}\and
Rupert Polley\inst{1,2}\and
Qingyuan Zhang\inst{1}\and
Yin Wu\inst{2,3}\and \\
Philip Schörner\inst{1}\and
Carl Esselborn\inst{3}\and
J. Marius Zöllner\inst{1,2} 
}

\authorrunning{A.~Abouelazm et al.}

\institute{FZI Research Center for Information Technology (\email{abouelazm@fzi.de})
\and Karlsruhe Institute of Technology \and CARIAD SE
}

\maketitle
\begin{abstract}
End-to-end planners based on waypoint regression achieve strong open-loop accuracy, but they primarily learn to mimic expert geometry and remain difficult to adapt to deployment-time safety constraints.
We propose a query-based cost-learning framework that estimates bounded costs for dynamically reachable ego trajectory queries, rather than dense BEV cells or a small regressed trajectory set.
Compact joint scene tokens capture coherent multimodal agent futures, while contingency-aware cost aggregation and cost-guided intra-cluster MPPI mixing convert the learned cost topology into feasible ego plans.
On nuScenes, our method improves over prior cost-estimation planners such as ST-P3 and NMP, outperforms most regression baselines in collision rate, while remaining competitive in L2, and retaining an interpretable cost interface.
On real-world driving logs, the proposed planner reduces collision rates compared with SparseDrive and Alpamayo without fine-tuning, while maintaining a diverse set of candidate trajectories.
\end{abstract}

\section{Introduction}
\label{sec:intro}

\EtoE (\eetoe) driving has progressed from direct sensor-to-control learning~\cite{hubschneider2017adding,bojarski2017explaining} to unified systems that jointly learn perception, prediction, and planning~\cite{chitta2022transfuser,hu2023planning}.
Recent sparse representations make these systems more scalable by allocating capacity to compact task- or instance-level tokens instead of dense raster features~\cite{jiang2023vad,sun2025sparsedrive,jia2025drivetransformer}.
However, the final planning interface remains a bottleneck: decisions should remain risk-aware, interpretable, physically feasible, and adaptable to deployment-time safety constraints.

Waypoint regression approaches are effective because they directly predict a short horizon of future ego motions. However, imitation objectives primarily reward spatial similarity to expert demonstrations and do not explicitly distinguish safe behavior from collision-causing hard negatives~\cite{wang2026beyond}.
Consequently, regression planners are difficult to inspect or repair, may produce infeasible behavior under distribution shift, and weakly incorporate safety through post-hoc rescoring of a limited top-$k$ set.
Cost maps offer a more abstract alternative by exposing a risk landscape that can be optimized, repaired, and combined with external costs at runtime.

Existing \eetoe cost-map planners such as Neural Motion Planner (NMP)~\cite{zeng2019end} and ST-P3~\cite{hu2022st}, however, estimate dense rasterized costs over many unreachable cells, spreading supervision away from feasible trajectories.
Their flexibility has therefore often come at the cost of lower open-loop planning accuracy than direct regression.
Moreover, many planners rely on marginal motion modes, which can ignore cross-agent correlations and miss scene-consistent interaction outcomes~\cite{sun2022m2i}.
This motivates a cost representation that preserves the optimization benefits of cost maps while avoiding dense supervision over irrelevant space. 

We propose a vectorized, query-based \eetoe cost planner that narrows this gap by learning a cost estimation over the reachable ego trajectories rather than over a dense spatial grid.
The model estimates mode-conditioned costs for sampled ego trajectory queries from compact joint scene tokens built from vectorized map and agent representations.
It then aggregates costs over coherent future scene modes with a contingency strategy and refines trajectories through cluster-based MPPI-style mixing guided by the learned cost topology.
This design retains the flexibility of cost optimization while approaching regression-style accuracy and improving transfer to real-world driving. The key contributions of this work are:
\begin{itemize}
    \item Query-based cost estimation over sampled reachable ego trajectories instead of dense raster grids.
    \item Compact joint scene tokens that condition ego costs on coherent multimodal futures for contingency planning.
    \item Cost-guided intra-cluster mixing for continuous reachable-space coverage beyond top-$k$ selection.
    \item Ranking-based cost learning objective driven by expert trajectories, collision risk, and off-road violations.
\end{itemize}
\begin{figure*}[t]
    \centering
    \includegraphics[
        width=0.97\linewidth,
        trim=20pt 7pt 12pt 3pt,
        clip
    ]
    {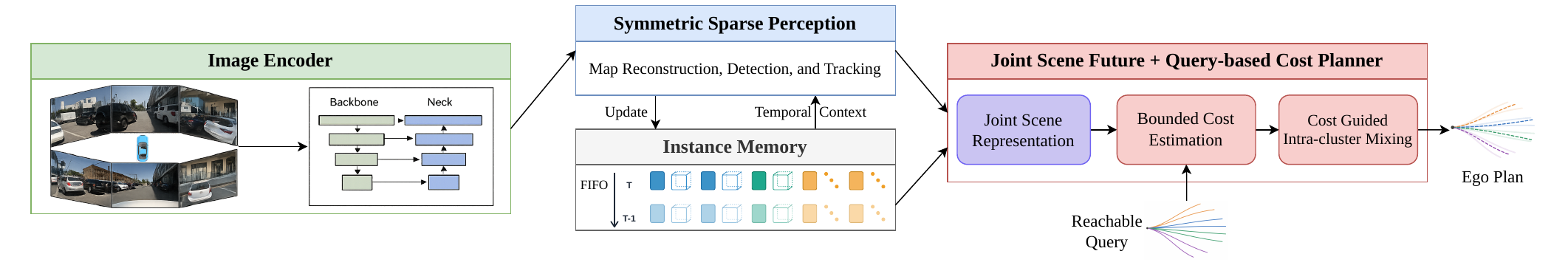}
    \caption{
        Multi-view images feed a sparse perception encoder with temporal instance memory. The proposed planner combines joint scene representations and reachable ego queries to estimate bounded costs and refine ego plans through intra-cluster mixing.
    }
    \label{fig:method_overview}
\end{figure*}
\section{Related Work}
\label{sec:sota}
\eetoe autonomous driving has advanced along two closely related directions: richer scene representations and more structured planning interfaces. Our work lies at their intersection, using compact sparse scene tokens while addressing the limitations of direct regression by estimating costs over reachable ego trajectories.

\subsection{Sparse and vectorized end-to-end planning.}

\eetoe driving has moved from intermediate-representation and sensor-fusion models, such as DSDNet~\cite{zeng2020dsdnet} and TransFuser~\cite{chitta2022transfuser}, toward planning-oriented architectures such as UniAD~\cite{hu2023planning} and DriveTransformer~\cite{jia2025drivetransformer}. In parallel, vectorized and sparse representations have reduced the computation cost of dense BEV representations. MapTR introduced structured queries for online map reconstruction~\cite{liao2022maptr}; VAD, SparseDrive, and SparseAD extend this idea to end-to-end driving with compact map and agent queries~\cite{jiang2023vad,sun2025sparsedrive,zhang2024sparsead}.
While these methods improve scene encoding and planning efficiency, the final decision is commonly obtained through trajectory regression. They also often reason over marginal agent modes, which can fail to capture coherent scene understanding and complex interactive futures~\cite{sun2022m2i}. In contrast, our planner learns a cost landscape over reachable ego trajectories conditioned on joint scene tokens, so each action is scored against coherent multimodal scene hypotheses rather than independently selected futures.

\subsection{Cost-based and safety-aware planning.}
Cost-based planners provide an interpretable interface between perception and action by exposing a cost landscape that can be optimized or repaired. Neural Motion Planner~\cite{zeng2019end}, QUAD~\cite{biswas2024quad}, and MP3~\cite{casas2021mp3} estimate costs from BEV or map-conditioned features and score sampled ego trajectories, while ST-P3 learns a spatial-temporal BEV cost volume for multi-camera planning~\cite{hu2022st}. These works demonstrate the value of learned costs, but dense BEV cost supervision disperses learning over large regions that are irrelevant to the ego vehicle's reachable motion. This can make optimization less stable, blur the distinction between reachable and unreachable costs, and has often led to weaker planning performance than direct regression.

In addition, cost learning is connected to \IRL (\irl), which treats expert demonstrations as evidence of an underlying reward or cost function. Maximum-entropy \irl formalizes this idea by recovering costs whose induced trajectory distribution matches expert behavior~\cite{wulfmeier2015maximum}, with later extensions to spatial-temporal costmaps~\cite{lee2022spatiotemporal,lee2022risk,triest2023learning}. 
However, these methods usually supervise costs indirectly through expert likelihood or policy matching, often requiring an inner planning or policy-inference loop.
On the other hand, purely supervised trajectory regression has safety limitations: Wang et al.~\cite{wang2026beyond} show that distance-to-expert losses fail to separate safe and collision-prone alternatives. We instead learn a query-based cost landscape in which expert behavior, collision risk, and off-road violations supervise the ordering of reachable ego queries.

\section{Methodology}
\label{sec:method}
\subsection{Overview and Problem Formulation}

Given a history of multi-view images
$\mathcal{I}=\{I_{t,n}\in\mathbb{R}^{3\times H_0\times W_0}\}_{t=-T_{\mathrm{hist}}}^{0}, n=1,\ldots,N_I$,
the current ego state $\mathbf{x}_0$, and a high-level navigation command $c$, 
the planner outputs a future ego trajectory
$\hat{\tau}=\{\hat{\mathbf{x}}_t\}_{t=1}^{T_{\mathrm{plan}}}$ in the ego coordinate frame.
Trajectory-regression planners learn this mapping directly, producing a small set of probabilistic ego futures.
In contrast, we formulate planning as cost estimation over dynamically reachable ego trajectory queries.
This preserves the interpretability of cost-based planning while avoiding dense supervision over regions the ego vehicle cannot execute.

Figure~\ref{fig:method_overview} summarizes the proposed architecture, which follows four stages.
First, a sparse perception encoder extracts compact agent and map tokens from the image queue.
Second, a joint scene prediction module summarizes multimodal agent futures into joint scene tokens
$\mathcal{S}=\{(\mathbf{s}_j,\pi_j)\}_{j=1}^{N_s}$, where $\mathbf{s}_j$ denotes a future scene-mode token, $\pi_j$ its probability, and $N_s$ the number of scene modes.
Third, conditioned on the ego state $\mathbf{x}_0$ and navigation command $c$, a kinematic rollout sampler generates a clustered reachable query set 
$\mathcal{Q}=\{(\tau_i,g_i)\}_{i=1}^{N}$,
where $\tau_i$ is a feasible ego trajectory and $g_i$ denotes an intention cluster induced by the sampler.
Each trajectory is then embedded as an ego query.
Finally, the model predicts a bounded cost for each ego query, scene mode, and future timestep,
\begin{equation}
    \mathbf{C}_{\theta}
    =
    \big[C_{\theta}(\tau_i,\mathbf{s}_j,t)\big]
    \in [C_{\min}, C_{\max}]^{N\times N_s\times T_{\mathrm{plan}}}.
    \label{eq:query_cost_tensor}
\end{equation}
where $\theta$ denotes the learned model parameters.
We refer to $\mathbf{C}_{\theta}$ as a query-based cost estimate: rather than assigning costs to dense BEV cells, it defines a cost landscape over the reachable ego trajectory manifold.
The final plan is obtained by aggregating costs over time and scene modes with a contingency-aware risk criterion, refining low-cost candidates through cost-guided trajectory mixing within intention clusters, and selecting the refined trajectory with the lowest resulting cost.

\subsection{Vectorized Map and Agent Encoding}

To efficiently encode the image queue $\mathcal{I}$ into compact map and agent features, we build on the SparseDrive image encoder and symmetric sparse perception module~\cite{sun2025sparsedrive}. 
At each timestep, a ResNet-FPN~\cite{lin2017feature} image encoder maps the multi-view images to multi-scale image features

\begin{equation}
    \mathcal{F}=\{\mathbf{F}_r\in\mathbb{R}^{N_I\times D\times H_r\times W_r}\}_{r=1}^{N_F},
    \label{eq:image_features}
\end{equation}

where $D$ is the feature dimension and $N_F$ is the number of feature scales. 
The sparse perception module then learns two distinct instance sets.
Static map elements are encoded as map features and polyline anchors $(\mathbf{M},\mathbf{L})$, 
with $\mathbf{M}\in\mathbb{R}^{N_m\times D}$ and $\mathbf{L}\in\mathbb{R}^{N_m\times P\times 2}$. 
Dynamic agents are encoded as agent features and 3D anchors $(\mathbf{A},\mathbf{B})$, 
with $\mathbf{A}\in\mathbb{R}^{N_a\times D}$ and $\mathbf{B}\in\mathbb{R}^{N_a\times 11}$.
Unlike BEV-based encoders~\cite{hu2022st}, SparseDrive does not lift image features into a dense BEV volume.
Instead, each sparse instance uses its anchor to gather image evidence from $\mathcal{F}$ through deformable feature aggregation.
After confidence-based selection, the map tokens $\mathbf{M}$ and agent tokens $\mathbf{A}$ provide the initial vectorized context for the joint prediction and cost estimation modules.

\subsection{Joint Scene Representation}

Starting from the selected map and agent tokens, temporal propagation and attention-based interactions refine the features with latent temporal and local scene context.
However, cost-based planning also requires compact hypotheses about how the scene may evolve.
Predicting agents independently can produce futures that are individually likely but mutually inconsistent~\cite{weng2023joint}.
Accordingly, we propose $N_s$ joint scene modes, each assigned a scene-level probability, so the cost estimator can reason over coherent scene contingencies rather than independent marginal futures.

Each scene mode is represented by a learnable embedding $\mathbf{e}_j\in\mathbb{R}^{D}$ shared across all agents.
We form mode-conditioned agent tokens by adding this embedding to each selected agent token,
$\tilde{\mathbf{a}}_{a,j}=\mathbf{a}_a+\mathbf{e}_j$,
yielding $\tilde{\mathbf{A}}\in\mathbb{R}^{N_a\times N_s\times D}$.
A scene-level token is produced by cross-attending the mode embedding to its mode-conditioned agent tokens:
\begin{equation}
    \mathbf{s}_j =
    \mathrm{FFN}\!\left(
    \mathrm{MHCA}
    \left(
    Q=\mathbf{e}_j,\;
    K=\{\tilde{\mathbf{a}}_{a,j}\}_{a=1}^{N_a},\;
    V=\{\tilde{\mathbf{a}}_{a,j}\}_{a=1}^{N_a}
    \right)
    \right).
    \label{eq:scene_attention}
\end{equation}

The resulting token $\mathbf{s}_j$ summarizes one coherent future scene mode, and its probability is predicted by
$\pi_j=\mathrm{softmax}_j(\psi_{\mathrm{scene}}(\mathbf{s}_j))$,
where $\psi_{\mathrm{scene}}$ is a lightweight scene-logit head.
To encourage these scene tokens to represent jointly consistent futures, the mode-conditioned agent tokens regress mode-specific agent trajectories, while the scene tokens predict the mixture weights over modes~\cite{shi2024mtr++}.
Let $\mathbf{Y}=\{\mathbf{y}_{a,t}\}$ denote the future positions of the selected agents for $t=1,\ldots,T_{\mathrm{pred}}$.
We model their joint future distribution as

\begin{equation}
    p(\mathbf{Y}\mid\mathcal{I})
    =
    \sum_{j=1}^{N_s}\pi_j
    \prod_{a=1}^{N_a}\prod_{t=1}^{T_{\mathrm{pred}}}
    \mathcal{N}\!\left(
    \mathbf{y}_{a,t};
    \boldsymbol{\mu}_{a,j,t},
    \boldsymbol{\Sigma}_{a,j,t}
    \right).
    \label{eq:joint_scene_gmm}
\end{equation}

The joint prediction branch is trained with the negative log-likelihood of Eq.~\ref{eq:joint_scene_gmm}, encouraging the scene-mode probabilities to assign mass to the hypotheses that best explain the observed joint future.

\subsection{Reachable Ego Queries and Cost Estimation}
\begin{figure}[tbp]
    \centering
    \includegraphics[
        width=0.96\linewidth,
        trim=30pt 14pt 50pt 10pt,
        clip
    ]
    {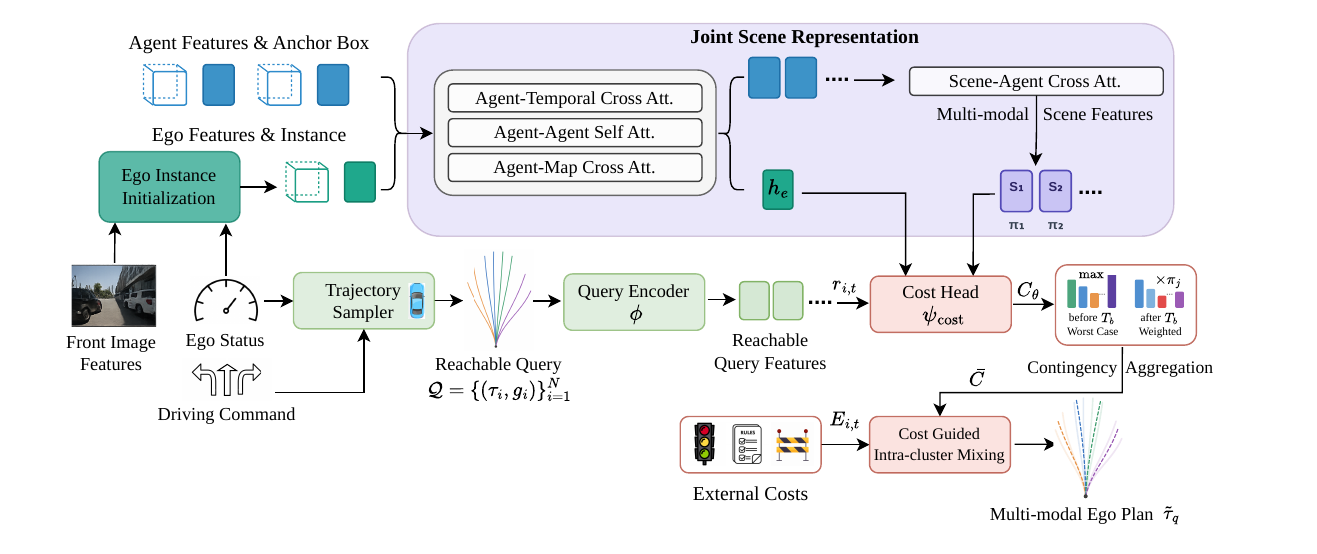}
    \caption{
        Detailed view of the proposed planner.
        Joint scene tokens encode multimodal futures, while the ego branch samples and encodes reachable trajectory queries.
        A lightweight head predicts scene-conditioned costs with contingency aggregation, optional inference-time external-cost repair, and cost-guided intra-cluster mixing.
    }
    \label{fig:method_details}
\end{figure}

Figure~\ref{fig:method_details} details how joint scene tokens and reachable ego queries are combined to estimate and refine query costs.
The previous modules provide a compact description of the current scene and its plausible joint futures.
We build on this representation to define the planning interface itself: instead of regressing a few ego motions, we learn costs over explicit reachable ego queries.
The ego branch builds a context token for the current decision by pooling an ego feature from the front-camera feature map and refining it through temporal attention, agent interaction, and cross-attention to selected map tokens. 
We then add a learned embedding of the navigation command $c$, yielding an ego context token $\mathbf{h}_e\in\mathbb{R}^{D}$ that encodes the current ego state, scene context, and route intent.

\subsubsection{Reachable Query Sampling.}
To avoid the diversity collapse often associated with top-$k$ selection, we generate reachable queries with intention-aware clustering.
Given the current ego state $\mathbf{x}_0$ and command $c$, a kinematic rollout sampler produces
$\mathcal{Q}=\{(\tau_i,g_i)\}_{i=1}^{N}$, where $\tau_i=\{\mathbf{x}_{i,t}\}_{t=1}^{T_{\mathrm{plan}}}$ is a dynamically feasible ego trajectory and $g_i\in\{1,\ldots,N_g\}$ denotes its intention cluster.
The clusters preserve coverage over distinct maneuver families, while the samples within each cluster explore local motion variations.
Each trajectory contributes its geometry and temporal structure through a coordinate and time embedding,
\begin{equation}
    \mathbf{r}_{i,t}
    =
    \phi_{\tau}\!\left(\mathrm{PE}(\mathbf{x}_{i,t})\right)
    +
    \boldsymbol{\gamma}_t,
    \label{eq:ego_query_embedding}
\end{equation}
where $\phi_{\tau}$ is a lightweight projection and $\boldsymbol{\gamma}_t$ denotes the timestep embedding.

For cost estimation, the three inputs play complementary roles: $\mathbf{h}_e$ provides the current ego-scene context, $\mathbf{r}_{i,t}$ provides the spatial location and timing of a candidate ego motion, and $\mathbf{s}_j$ provides a compact hypothesis of the surrounding scene future.
We combine them additively to form a cost query for trajectory $i$, scene mode $j$, and timestep $t$:
\begin{equation}
    \mathbf{z}^{\mathrm{cost}}_{i,j,t}
    =
    \mathbf{r}_{i,t}
    +
    \mathbf{h}_{e}
    +
    \mathbf{s}_{j}.
    \label{eq:cost_query_token}
\end{equation}
A lightweight cost head maps these queries to bounded costs,
\begin{equation}
    C_{\theta}(\tau_i,\mathbf{s}_j,t)
    =
    C_{\min}
    +
    \sigma\!\left(\psi_{\mathrm{cost}}(\mathbf{z}^{\mathrm{cost}}_{i,j,t})\right)
    (C_{\max}-C_{\min}),
    \label{eq:bounded_query_cost}
\end{equation}
where $\psi_{\mathrm{cost}}$ is implemented as a small prediction head shared across trajectory queries.
This defines a query-based cost map only over executable ego trajectories, avoiding dense BEV cells unreachable by the vehicle.

\subsubsection{Risk-aware Aggregation.}
The per-mode costs must be collapsed without prematurely committing to a single future scene.
This is important in interactive settings, where early ego actions should remain safe under multiple plausible agent responses, while later costs can account for the probability of each contingency~\cite{li2025multimodal}.
We therefore use a risk-aware contingency aggregation.
Let $b=\lfloor T_b/\Delta t \rfloor$ denote the branching index, where $T_b$ is the branching time in seconds and $\Delta t$ is the planning timestep.
Before this index, the planner uses the worst scene-mode cost; afterward, it follows the learned scene probabilities:
\begin{equation}
    \bar{C}_{i,t}
    =
    \begin{cases}
    \max_j C_{\theta}(\tau_i,\mathbf{s}_j,t), & t \leq b,\\
    \sum_{j=1}^{N_s}\pi_j \, C_{\theta}(\tau_i,\mathbf{s}_j,t), & t > b.
    \end{cases}
    \label{eq:contingency_cost}
\end{equation}

\subsubsection{Cost-guided Intra-cluster Mixing.}
The aggregated costs define a local cost topology over the sampled reachable set.
Pure sampling provides broad coverage, but the hard selection remains limited by the resolution of the sampled set and can therefore suffer from high displacement error.
Selecting only the global lowest-cost samples can also collapse to a single maneuver family.
We address these limitations with cost-guided intra-cluster mixing: intention clusters preserve global maneuver diversity, while the learned risk-aware costs refine each cluster toward its low-cost region.

Let $\mathcal{J}_q=\{i:g_i=q\}$ denote the samples in intention cluster $q$.
For mixing, we use a flexible cost profile $R_{i,t}$ derived from $\bar{C}_{i,t}$.
This profile uses average trajectory cost by default, with per-timestep
cost as an alternative.
Within each cluster, this profile is converted into MPPI-style weights~\cite{williams2017model},
\begin{equation}
    w_{i,t}^{(q)}
    =
    \frac{
    \exp\!\left(-(R_{i,t} - R_{q,t}^{\min})/\lambda\right)}
    {\sum_{m\in\mathcal{J}_q}
    \exp\!\left(-(R_{m,t} - R_{q,t}^{\min})/\lambda\right)},
    \qquad
    R_{q,t}^{\min}=\min_{m\in\mathcal{J}_q}R_{m,t},
    \label{eq:mppi_cluster_weights}
\end{equation}
where $\lambda$ controls how strongly the refinement follows the lowest-cost samples.
Given the sampled control sequence $\mathbf{u}_{i,t}$ associated with each trajectory, the refined control for cluster $q$ is computed directly as
\begin{equation}
    \tilde{\mathbf{u}}_{q,t}
    =
    \sum_{i\in\mathcal{J}_q} w_{i,t}^{(q)}\mathbf{u}_{i,t},
    \qquad
    \tilde{\tau}_q=f_{\mathrm{kin}}(\{\tilde{\mathbf{u}}_{q,t}\}_{t=1}^{T_{\mathrm{plan}}}),
    \label{eq:mppi_cluster_fusion}
\end{equation}
where $f_{\mathrm{kin}}$ denotes the ego kinematic rollout.
The resulting set $\{\tilde{\tau}_q\}$ retains one candidate per intention 
while using the learned safety-aware costs to refine controls beyond the original samples.
The same weights define a cluster-desirability proxy for the mixed-trajectory cost,
\begin{equation}
    \tilde{C}_q
    =
    \frac{1}{T_{\mathrm{plan}}}\sum_{t=1}^{T_{\mathrm{plan}}}
    \sum_{i\in\mathcal{J}_q}w_{i,t}^{(q)}R_{i,t}.
    \label{eq:mppi_cluster_cost}
\end{equation}
This proxy preserves a single differentiable path through which trajectory ranking, classification, and regression shape sample costs and fusion, without rescoring the mixed trajectory.
For final candidate selection and classification supervision, we use the negative cluster costs as logits,
$p_q^{\mathrm{ego}}=\mathrm{softmax}_q(-\tilde{C}_q)$, such that lower-cost refined candidates receive higher probability.

\subsection{Cost Ranking and Candidate Supervision}

During training, we retain the standard SparseDrive losses for map and object detection and supervise the joint scene modes with the negative log-likelihood in Eq.~\ref{eq:joint_scene_gmm}.
We then add a planner-specific objective for the query-based cost estimation.
This planning objective supervises two complementary parts of the planner: the cost topology over reachable queries and the final candidate selected after intra-cluster refinement. 
We propose the following planning loss as
\begin{equation}
    \mathcal{L}_{\mathrm{plan}}
    =
    \lambda_{\mathrm{rank}} \, \mathcal{L}_{\mathrm{rank}}
    +
    \lambda_{\mathrm{cls}}  \, \mathcal{L}_{\mathrm{cls}}
    +
    \lambda_{\mathrm{reg}} \, \mathcal{L}_{\mathrm{reg}}.
    \label{eq:planning_loss}
\end{equation}
Here, $\mathcal{L}_{\mathrm{rank}}$ trains the cost map before final selection, while $\mathcal{L}_{\mathrm{cls}}$ and $\mathcal{L}_{\mathrm{reg}}$ supervise the refined candidates $\{\tilde{\tau}_q\}_{q=1}^{N_g}$.

\subsubsection{Safety-aware cost ranking.}
Imitation losses based only on expert distance may assign similar supervision to safe alternatives and unsafe hard negatives~\cite{wang2026beyond}. 
We therefore train the learned costs to separate the expert trajectory $\tau^\star$ from sampled reachable trajectories using a novel safety-aware margin.
The expert trajectory is queried only during training, not at inference.
Unlike ST-P3- or NMP-style objectives~\cite{hu2022st,zeng2019end}, which mainly rank candidates through expert distance, our margin uses collision and boundary violations to determine the required separation between learned costs. 
For a sampled trajectory $\tau_i$, let $C^\star$ and $C_i$ denote the predicted costs of the expert and sampled trajectory under the chosen cost aggregation. 
The ranking violation and loss are
\begin{equation}
    v_i
    =
    \left[
    C^\star - C_i + m_i
    \right]_+,
    \qquad
    \mathcal{L}_{\mathrm{rank}}
    =
    \frac{1}{|\mathcal{K}_h|}
    \sum_{i\in\mathcal{K}_h} v_i .
    \label{eq:safety_rank_loss}
\end{equation}
where $m_i=m(d_i,\rho_i^{\mathrm{col}},\rho_i^{\mathrm{bd}})$ is a bounded margin that grows with the distance $d_i$ to the expert and increases sharply when $\tau_i$ collides or leaves the drivable boundary.
Intuitively, the loss enforces $C_i \geq C^\star + m_i$: candidates farther from the expert require a higher cost, and collision or boundary violations increase the margin even when the trajectory remains geometrically close to the expert.
Rather than using only the single largest violation, $\mathcal{K}_h$ denotes the indices of the top-$k_h$ violations in a scene.
This spreads gradients over informative negatives, producing a smoother topology and providing more stable guidance for the cost-weighted cluster mixing.

\subsubsection{Candidate selection and refinement.}
After intra-cluster mixing, each intention cluster yields one refined candidate $\tilde{\tau}_q$. 
The classification loss encourages the cluster-cost logits to select the candidate closest to the expert, while the regression loss pulls the selected candidate toward the expert trajectory. 
Together, the ranking loss shapes the intra-cluster cost landscape used for mixing, and the classification and regression losses preserve accurate inter-cluster selection while allowing the refined candidates to continuously cover the expert distribution.

\subsection{Inference-time Cost Repair}
\label{subsec:Inference-time Cost Repair}
At inference time, we propose a repair mechanism that incorporates external safety or rule signals without retraining the model.
Since our planner predicts bounded costs over explicit ego queries, these signals can directly override unsafe regions of the learned cost landscape.
Let $E_{i,t}\in[0,1]$ denote an external penalty for trajectory $\tau_i$ at timestep $t$, obtained, for example, from a traffic-rule monitor or vehicle-to-everything (V2X) traffic-light data. 
We perform cost repair by saturating externally unsafe query states toward the maximum cost,
\begin{equation}
    \bar{C}^{\, \mathrm{repair}}_{i,t}
    =
    (1-E_{i,t})\, \bar{C}_{i,t}
    +
    E_{i,t} \, C_{\max}.
    \label{eq:cost_repair}
\end{equation}
The repaired costs are then used in the same cluster-wise mixing described above.
Thus, trajectories marked unsafe by external signals receive low mixing weights, and intention clusters whose trajectories are consistently penalized obtain high cluster costs and are unlikely to be selected.
This gives the planner a practical repair mechanism for correcting learned costs at deployment while preserving the reachable-query structure.

\section{Experimental Setup}
\label{sec:setup}
\subsubsection{Dataset and protocol.}
We evaluate on the nuScenes dataset~\cite{caesar2020nuscenes}, which contains 1000 driving scenes with six surround-view cameras and keyframe annotations at 2Hz.
Using the train/validation split, the model observes a queue of four multi-view frames and produces a 3s future ego plan for open-loop evaluation. 
For zero-shot transfer, we use a subset of 465 driving logs from FZI-AURA~\cite{polley_fzi_aura_2026}, each lasting 20s, totaling 2.58 hours of driving, about 3$\times$ the duration of the nuScenes validation split. The logs were collected with the CoCar NextGen research vehicle~\cite{heinrich2024cocar} in Germany across different times of day, lighting conditions, and traffic densities. To reduce domain shift, we select six surround-view cameras and process the logs following a nuScenes-inspired protocol with manually annotated 3D bounding boxes and semantic segmentation labels. The vehicle uses a sensor rig similar in layout but not identical to nuScenes, resulting in slightly different camera intrinsics and extrinsics.


\subsubsection{Implementation details.}
Our model uses SparseDrive-S~\cite{sun2025sparsedrive} as the sparse perception backbone with a ResNet-50 FPN image encoder, 900 agent anchors, 100 map anchors, and 256-dimensional instance features. All variants of our model are trained with AdamW using a learning rate of $3\times10^{-4}$, weight decay of $10^{-3}$, and an effective batch size of 48 on two NVIDIA RTX 6000 Ada GPUs.
The query-cost planner uses $N_s=6$ joint scene modes, $N=512$ sampled ego trajectories, and $N_g=16$ intention clusters with 32 samples per cluster.
Costs are bounded to $[0,100]$, contingency aggregation uses a $T_b$ of 1s, and the safety-ranking loss averages the top-$k_h=32$ violations.

\subsubsection{Metrics and baselines.}
We report displacement error and collision rate at 1s, 2s, and 3s, together with their averages.
For zero-shot real-world transfer, we additionally report candidate diversity for all models.
Diversity is measured by the mean pairwise average distance and endpoint distance among candidate trajectories, capturing separation between maneuver intentions.
For variants of our end-to-end model, we also report cost AUC and cost gap, which measure how well the learned costs separate unsafe candidates from safe ones.
Unsafe candidates are those that collide or leave the drivable area.
An AUC of $0.5$ corresponds to random ordering and $1.0$ to perfect separation.
The cost gap is the difference between the mean cost of unsafe candidates and the mean cost of safe candidates; larger positive values indicate stronger safety-aligned separation.

\section{Evaluation}
\label{sec:eval}
\subsection{Comparison to State of the Art}

Table~\ref{tab:sota_planning} compares open-loop planning performance on nuScenes.
Compared with prior cost-based planners, our method substantially improves both displacement error and collision rate.
Our query cost representation concentrates learning on reachable ego motions, the safety-margin loss separates unsafe hard negatives, and cost-guided cluster mixing converts the learned topology into accurate final plans.
Against ST-P3, the average L2 error drops from 2.11m to 0.71m, while the average collision rate drops from 0.71\% to 0.07\%.
This closes much of the gap to recent regression-based planners while preserving the advantages of an explicit cost interface.
Notably, our method achieves a lower average collision rate than SparseDrive-S and remains close to the larger SparseDrive-B variant, while also outperforming most regression baselines such as VAD~\cite{jiang2023vad} and DriveTransformer~\cite{jia2025drivetransformer}, 
suggesting that learning costs over reachable queries provides a useful balance between accurate imitation and safety-aware planning.
\begin{table}[ht]
    \centering
    \caption{Open-loop planning performance on nuScenes under the UniAD metrics. $\dagger$ denotes LiDAR-based methods.}
    \label{tab:sota_planning}
    \resizebox{0.7\linewidth}{!}{%
    \setlength{\tabcolsep}{3.2pt}%
    \begin{tabular}{@{}lccc>{\columncolor[gray]{0.88}}cccc>{\columncolor[gray]{0.88}}c@{}}
        \toprule
        \multirow{2}{*}{Method} &
        \multicolumn{4}{c}{L2 (m) $\downarrow$} &
        \multicolumn{4}{c}{Collision (\%) $\downarrow$} \\
        \cmidrule(lr){2-5}\cmidrule(lr){6-9}
        & 1s & 2s & 3s & Avg. & 1s & 2s & 3s & Avg. \\
        \midrule
        \multicolumn{9}{@{}>{\columncolor[gray]{0.94}}c@{}}{\textbf{Cost-based methods}} \\
        \midrule
        NMP~\cite{zeng2019end}$^{\dagger}$        & --   & --   & 2.31 & --   & --   & --   & 1.92 & --   \\
        SA-NMP~\cite{zeng2019end}$^{\dagger}$     & --   & --   & 2.05 & --   & --   & --   & 1.59 & --   \\
        ST-P3~\cite{hu2022st}                     & 1.33 & 2.11 & 2.90 & 2.11 & 0.23 & 0.62 & 1.27 & 0.71 \\
        \midrule
        \textbf{Ours}                     & 0.28 & 0.65 & 1.21 & 0.71 & 0.00 & 0.02 & 0.18 & \underline{0.07} \\    
        \midrule
        \multicolumn{9}{@{}>{\columncolor[gray]{0.94}}c@{}}{\textbf{Regression-based methods}} \\
        \midrule
        FF~\cite{hu2021safe}$^{\dagger}$          & 0.55 & 1.20 & 2.54 & 1.43 & 0.06 & 0.17 & 1.07 & 0.43 \\
        EO~\cite{khurana2022differentiable}$^{\dagger}$     & 0.67 & 1.36 & 2.78 & 1.60 & 0.04 & 0.09 & 0.88 & 0.33 \\
        UniAD~\cite{hu2023planning}               & 0.48 & 0.74 & 1.07 & 0.76 & 0.12 & 0.13 & 0.28 & 0.17 \\
        VAD-Tiny~\cite{jiang2023vad}              & 0.46 & 0.76 & 1.12 & 0.78 & 0.21 & 0.35 & 0.58 & 0.38 \\
        VAD-Base~\cite{jiang2023vad}              & 0.41 & 0.70 & 1.05 & 0.72 & 0.07 & 0.17 & 0.41 & 0.22 \\
        BEVPlanner~\cite{li2024ego}            & 0.27 & 0.54 & 0.90 & 0.57 & --   & --   & --   & --   \\
        DriveTransformer~\cite{jia2025drivetransformer}
                                                        & 0.19 & 0.34 & 0.66 & 0.40 & 0.03 & 0.10 & 0.21 & 0.11 \\
        SparseDrive-S~\cite{sun2025sparsedrive}   & 0.29 & 0.58 & 0.96 & 0.61 & 0.01 & 0.05 & 0.18 & 0.08 \\
        SparseDrive-B~\cite{sun2025sparsedrive}   & 0.29 & 0.55 & 0.91 & 0.58 & 0.01 & 0.02 & 0.13 & \textbf{0.06} \\
        \bottomrule
    \end{tabular}
    }
\end{table}

\subsection{Zero-shot Real-world Transfer}

Table~\ref{tab:real_world_transfer} evaluates zero-shot transfer to real-world driving logs collected in Germany, without fine-tuning any method on this data.
Alpamayo~1.5~\cite{wang2025alpamayo} is a large vision-language driving model trained on substantially broader data, so we use it as a strong imitation-oriented reference rather than a lightweight planner baseline.
Compared with SparseDrive-S, our method reduces the average L2 distance from 4.22m to 2.31m and the average collision rate from 2.73\% to 0.68\%.
Alpamayo achieves the lowest displacement error, but uses a 10B-parameter model and requires 1.87s per frame on an NVIDIA RTX 3090 GPU.
In contrast, our 86.2M-parameter planner achieves the lowest collision rate, runs in real time, and maintains higher candidate diversity for downstream repair and selection. These results suggest that reachable-query cost learning provides a favorable balance between safety, diversity, and efficiency under real-world domain shift.
%

\begin{table}[tbp]
    \centering
    \caption{Zero-shot real-world transfer. No method is fine-tuned on the real-world logs. Latency is measured on a workstation with a RTX 3090 GPU.
    }
    \label{tab:real_world_transfer}
    \resizebox{\linewidth}{!}{%
    \setlength{\tabcolsep}{3.2pt}%
    \begin{tabular}{@{}lccc>{\columncolor[gray]{0.88}}cccc>{\columncolor[gray]{0.88}}ccccc@{}}
        \toprule
        \multirow{2}{*}{Method} &
        \multicolumn{4}{c}{L2 (m) $\downarrow$} &
        \multicolumn{4}{c}{Collision (\%) $\downarrow$} &
        \multicolumn{2}{c}{Diversity (m) $\uparrow$} &
        \multirow{2}{*}{Params.} &
        \multirow{2}{*}{\begin{tabular}{c}Latency\\(ms) $\downarrow$\end{tabular}} \\
        \cmidrule(lr){2-5}\cmidrule(lr){6-9}\cmidrule(lr){10-11}
        & 1s & 2s & 3s & Avg.
        & 1s & 2s & 3s & Avg.
        & Avg. & Endpoint
        & & \\
        \midrule
        SparseDrive-S~\cite{sun2025sparsedrive} & 2.38 & 4.17 & 6.11 & 4.22 & 0.53 & 2.57 & 5.08 & 2.73 & \underline{1.77} & \underline{3.41} & 85.9M & \underline{115} \\
        Alpamayo~1.5~\cite{wang2025alpamayo} & 0.64 & 1.41 & 2.39 & \textbf{1.48} & 0.03 & 0.90 & 1.79 & \underline{0.87} & 1.47 & 3.25 & 10.0B & 1878 \\
        \textbf{Ours}                           & 0.85 & 2.12 & 3.98 & \underline{2.31} & 0.01 & 0.40 & 1.50 & \textbf{0.68} & \textbf{5.54} & \textbf{11.88} & 86.2M & \textbf{104} \\
        \bottomrule
    \end{tabular}
    }
\end{table}



\subsection{Safety Supervision and Cost Topology}

We next isolate how the safety margin shapes the learned cost topology.
Table~\ref{tab:collision_supervision_ablation} first studies how the collision term in the margin is computed during training.
All variants use the same architecture and cost planner at inference; only the collision signal used to compute the training margin changes.
At test time, the margin itself is not required, since its effect is absorbed into the learned cost topology.
Ground-truth collision provides an oracle target, while top-1 predicted collision supervises the margin using only the most likely future scene.
In contrast, contingency collision risk is computed over multimodal predicted scene futures, allowing lower-probability but safety-critical futures to influence the learned costs.

The results show a tradeoff between accuracy and safety.
Top-1 predicted collision achieves the lowest displacement error, suggesting that supervising with the most likely predicted future is well aligned with imitation accuracy, but it also gives the highest average collision rate.
Using predicted contingency risk slightly increases L2 error, but reduces the average collision rate from 0.12\% to 0.07\%.
This indicates that multimodal collision supervision encourages the latent scene tokens to encode collision-relevant futures and yields a cost topology that is better aligned with safety-aware planning.

\begin{table}[tbp]
    \centering
    \caption{Ablation of collision supervision for the safety margin on nuScenes.}
    \label{tab:collision_supervision_ablation}
    \resizebox{0.7\linewidth}{!}{%
    \setlength{\tabcolsep}{3.2pt}%
    \begin{tabular}{@{}lccc>{\columncolor[gray]{0.88}}cccc>{\columncolor[gray]{0.88}}c@{}}
        \toprule
        \multirow{2}{*}{Collision supervision} &
        \multicolumn{4}{c}{L2 (m) $\downarrow$} &
        \multicolumn{4}{c}{Collision (\%) $\downarrow$} \\
        \cmidrule(lr){2-5}\cmidrule(lr){6-9}
        & 1s & 2s & 3s & Avg. & 1s & 2s & 3s & Avg. \\
        \midrule
        Ground-truth collision                & 0.27 & 0.63 & 1.16 & 0.69 & 0.00 & 0.06 & 0.25 & \underline{0.10} \\
        Top-1 predicted collision             & 0.25 & 0.59 & 1.10 & 0.64 & 0.02 & 0.07 & 0.26 & 0.12 \\
        \textbf{Contingency collision risk}   & 0.28 & 0.65 & 1.21 & 0.71 & 0.00 & 0.02 & 0.18 & \textbf{0.07} \\
        \bottomrule
    \end{tabular}
    }
\end{table}

Having chosen contingency collision risk as the default supervision, Tables~\ref{tab:cost_loss_ablation} and~\ref{tab:margin_source_ablation} evaluate whether the resulting cost topology is specific to the proposed ranking objective.
All variants use the same architecture, reachable-query planner, candidate losses, and inference procedure; only $\mathcal{L}_{\mathrm{rank}}$ is replaced.
We first compare against ranking losses adapted from NMP~\cite{zeng2019end} and ST-P3~\cite{hu2022st}, and then ablate the sources used to form our safety margin.

Table~\ref{tab:cost_loss_ablation} shows that the choice of cost supervision mainly changes the safety structure of the learned topology.
The ST-P3-style loss achieves the lowest nuScenes L2 loss, but its weaker cost separation results in a higher collision rate than the proposed margin.
In contrast, our safety-margin objective improves nuScenes collision from 0.12\% to 0.07\% relative to the ST-P3-style loss, while increasing cost AUC from 0.79 to 0.86 and cost gap from 7.80 to 45.70.
The same trend holds under real-world transfer: the proposed objective yields the lowest collision rate and the largest cost gap, indicating that the learned safety ordering generalizes beyond the training domain.
\begin{table}[ht]
    \centering
    \caption{Comparison of cost-supervision objectives. 
    }
    \label{tab:cost_loss_ablation}
    \resizebox{0.95\linewidth}{!}{%
    \setlength{\tabcolsep}{3.2pt}%
    \begin{tabular}{@{}lcccc|cccc@{}}
        \toprule
        \multirow{2}{*}{Objective $\mathcal{L}_{\mathrm{rank}}$} &
        \multicolumn{4}{c|}{nuScenes} &
        \multicolumn{4}{c}{Real-world} \\
        \cmidrule(lr){2-5}\cmidrule(lr){6-9}
        & L2 $\downarrow$ & Coll. $\downarrow$ & Cost AUC $\uparrow$ & Cost Gap $\uparrow$
        & L2 $\downarrow$ & Coll. $\downarrow$ & Cost AUC $\uparrow$ & Cost Gap $\uparrow$ \\
        \midrule
        NMP-style loss~\cite{zeng2019end}       & 0.76 & 0.33 & 0.70 & 6.60 & 2.29 & 0.84 & 0.59 & 3.99 \\
        ST-P3-style loss~\cite{hu2022st}        & 0.61 & 0.12 & 0.79 & 7.80 & 3.26 & 0.83 & 0.61 & 4.24 \\
        \textbf{Proposed safety margin}         & 0.71 & 0.07 & 0.86 & 45.70 & 2.31 & 0.68 & 0.63 & 16.18 \\
        \bottomrule
    \end{tabular}
    }
\end{table}

Table~\ref{tab:margin_source_ablation} further decomposes the proposed margin.
This evaluates the contribution of expert-distance and safety terms, and whether combining them yields a cost topology that is both safety-aware and useful for final planning. The distance-only margin keeps the trajectories close to the expert but leaves many unsafe hard negatives weakly separated, resulting in higher collision rates on both nuScenes and real-world logs.
Using only collision and off-road terms produces the strongest raw separation, but it also degrades displacement and does not translate into the best final collision rate.
Combining distance with collision and off-road terms gives the best planning tradeoff: the learned costs remain anchored to expert-like reachable motions while still penalizing unsafe alternatives, yielding the lowest collision rate in both domains.
\begin{table}[tbp]
    \centering
    \caption{Ablation of safety-margin sources.}
    \label{tab:margin_source_ablation}
    \resizebox{0.95\linewidth}{!}{%
    \setlength{\tabcolsep}{3.2pt}%
    \begin{tabular}{@{}lcccc|cccc@{}}
        \toprule
        \multirow{2}{*}{Margin source} &
        \multicolumn{4}{c|}{nuScenes} &
        \multicolumn{4}{c}{Real-world} \\
        \cmidrule(lr){2-5}\cmidrule(lr){6-9}
        & L2 $\downarrow$ & Coll. $\downarrow$ & Cost AUC $\uparrow$ & Cost Gap $\uparrow$
        & L2 $\downarrow$ & Coll. $\downarrow$ & Cost AUC $\uparrow$ & Cost Gap $\uparrow$ \\
        \midrule
        Distance                              & 0.72 & 0.20 & 0.71 & 10.51 & 2.36 & 1.71 & 0.56 & 4.89 \\
        Collision + off-road                  & 0.88 & 0.11 & 0.87 & 51.25 & 2.34 & 1.48 & 0.66 & 23.57 \\
        \textbf{All}  & 0.71 & 0.07 & 0.86 & 45.70 & 2.31 & 0.68 & 0.63 & 16.18 \\
        \bottomrule
    \end{tabular}
    }
\end{table}

\subsection{Qualitative Results of External Cost Repair}

\begin{figure*}[t]
    \centering
    \begin{subfigure}[t]{0.44\textwidth} 
        \centering 
        \includegraphics[ 
            width=0.99\linewidth, 
            trim={0 43bp 0 43bp}, clip 
        ]{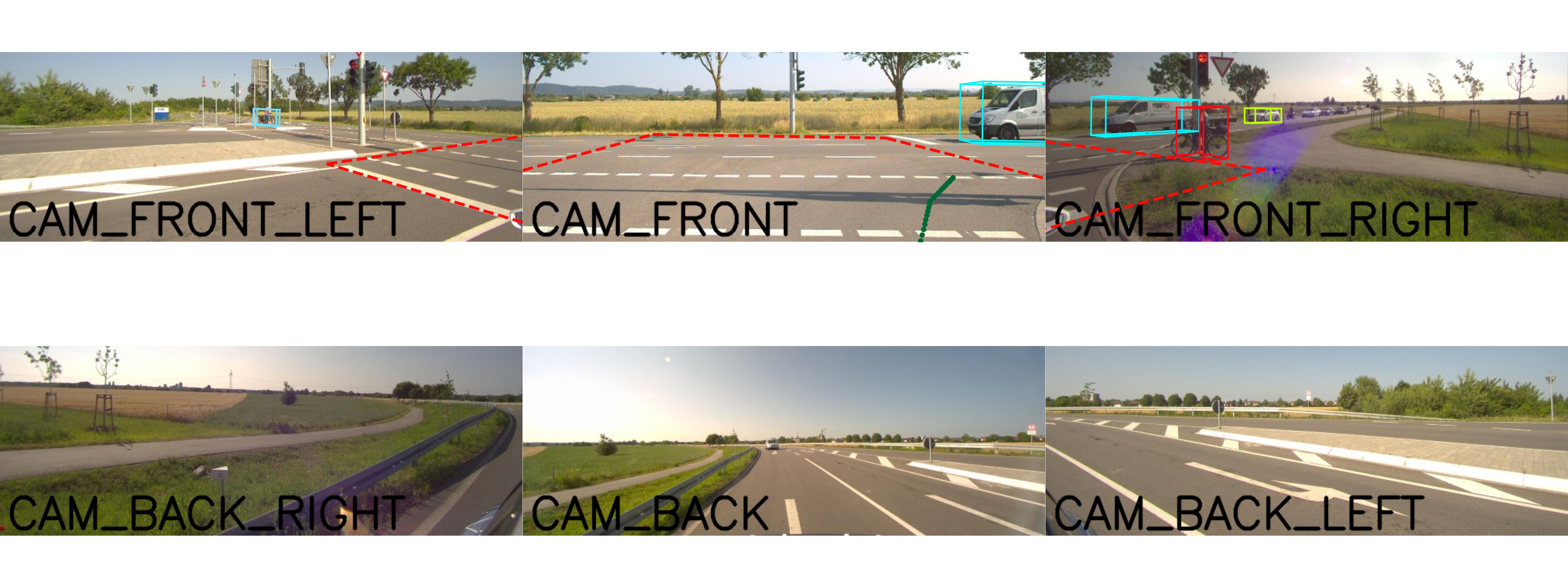} 
        \caption{Camera view} 
        \label{fig:external_cost_repair_base_cam} 
    \end{subfigure}
    \hfill
    \begin{subfigure}[t]{0.23\textwidth}
        \centering
        \includegraphics[
            width=0.97\linewidth,
            trim={200bp 800bp 0 0},
            clip
        ]{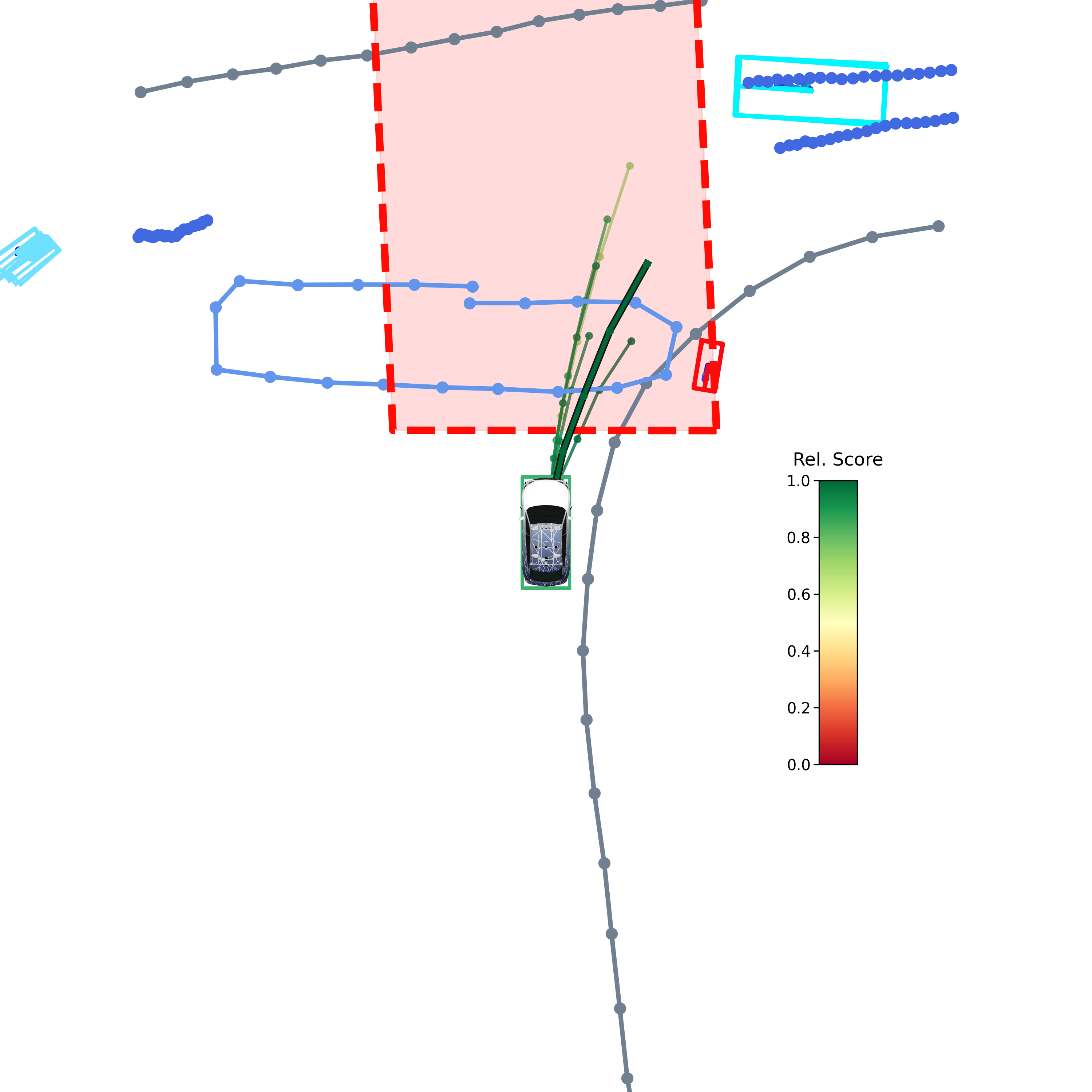}
        \caption{SparseDrive BEV}
        \label{fig:external_cost_repair_base_bev}
    \end{subfigure}
    \hfill
    \begin{subfigure}[t]{0.23\textwidth}
        \centering
        \includegraphics[
            width=0.97\linewidth,
            trim={200bp 800bp 0 0},
            clip
        ]{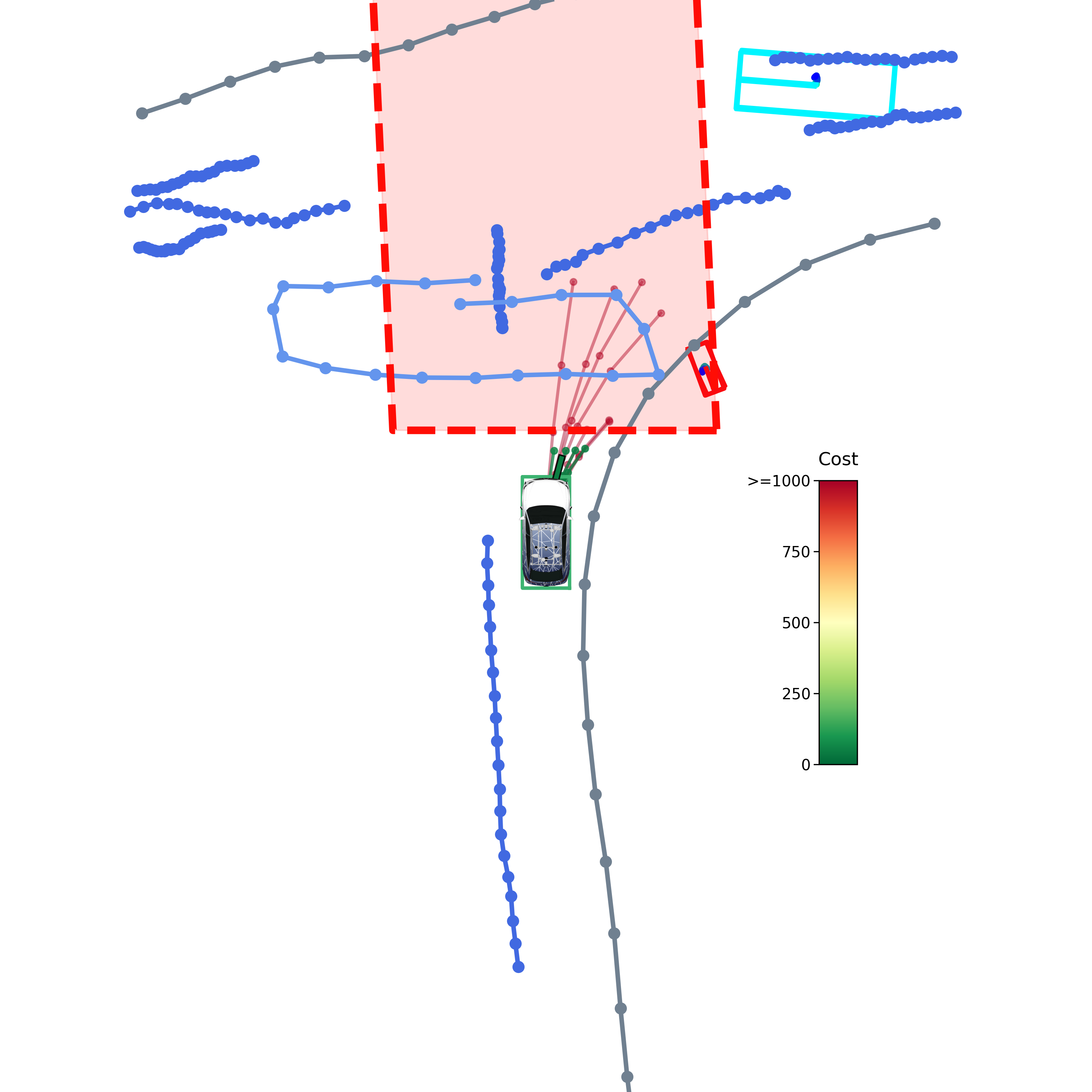}
        \caption{Ours BEV}
        \label{fig:external_cost_repair_ours_bev}
    \end{subfigure}
    \caption{
        Qualitative comparison using the conflict area of a red traffic light received from a V2X unit.
        SparseDrive's low-diversity proposals already violate the rule, so post-hoc rescoring cannot recover a rule-compliant trajectory.
        Our planner applies the external penalty to the aggregated query-cost profile, producing safe low-cost candidates.
    }
    \label{fig:external_cost_repair}
\end{figure*}

Figure~\ref{fig:external_cost_repair} illustrates the inference-time cost repair introduced in Section~\ref{subsec:Inference-time Cost Repair}.
As an example of external costs, we use the conflict area of a red traffic light received from V2X infrastructure.
SparseDrive produces a low-diversity proposal set in which all candidates enter this area (Fig.~\ref{fig:external_cost_repair_base_bev}).
Since the proposals already violate the rule, post-hoc rescoring alone cannot recover a rule-compliant trajectory.
Our planner instead applies the external penalty to the aggregated query-cost profile before intra-cluster mixing.
This changes the cost topology used to form the final candidates, suppressing unsafe queries and producing safe low-cost candidates, highlighted in green in Fig.~\ref{fig:external_cost_repair_ours_bev}, that respect the traffic-light rule.

\subsection{Qualitative Robustness to Misaligned Navigation Commands}
\label{subsec:wrong_command}

\begin{figure*}[t]
    \centering
    \newlength{\bevpanelheight}
    \setlength{\bevpanelheight}{0.34\textwidth}
    \begin{subfigure}[t]{0.44\textwidth}
        \centering
            \includegraphics[
                width=0.99\linewidth,
                trim={0 43bp 0 43bp},
                clip
            ]{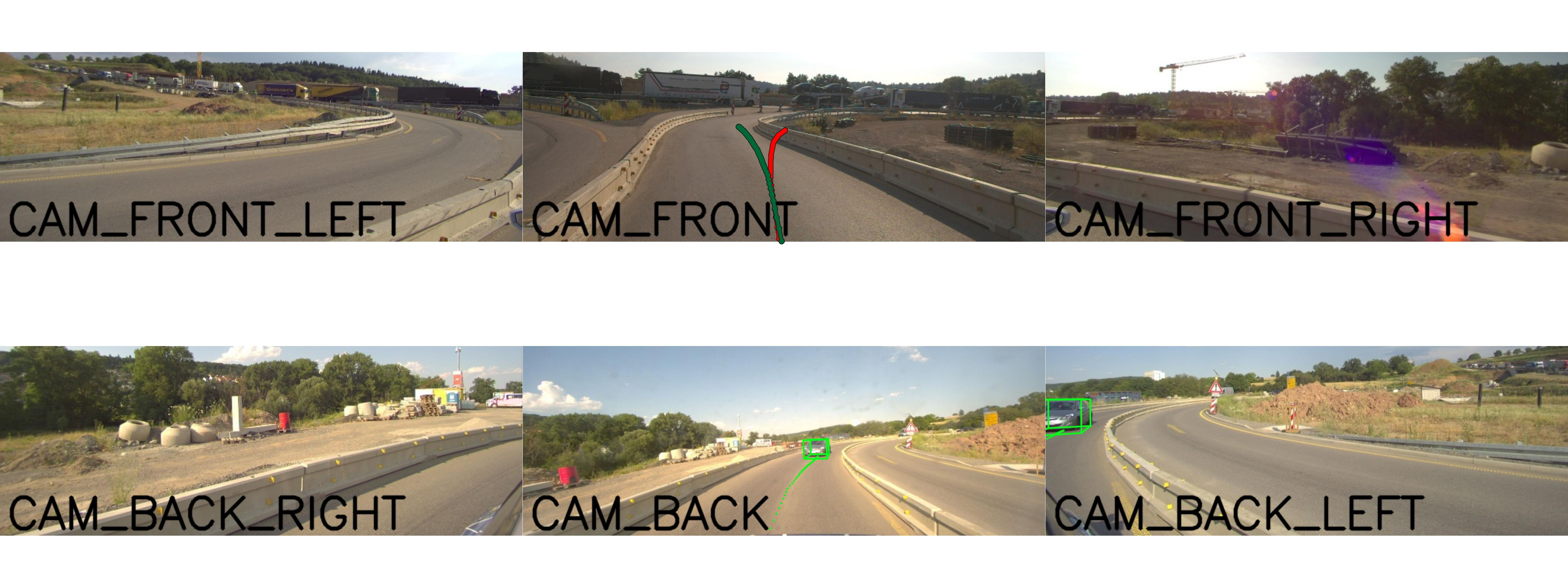}
        \caption{Projected final trajectory candidates}
        \label{fig:wrong_command_camera}
    \end{subfigure}
    \hfill
    \begin{subfigure}[t]{0.23\textwidth}
        \centering
            \includegraphics[
                height=0.97\linewidth,
                trim={800bp 1245bp 300bp 120bp},
                clip
            ]{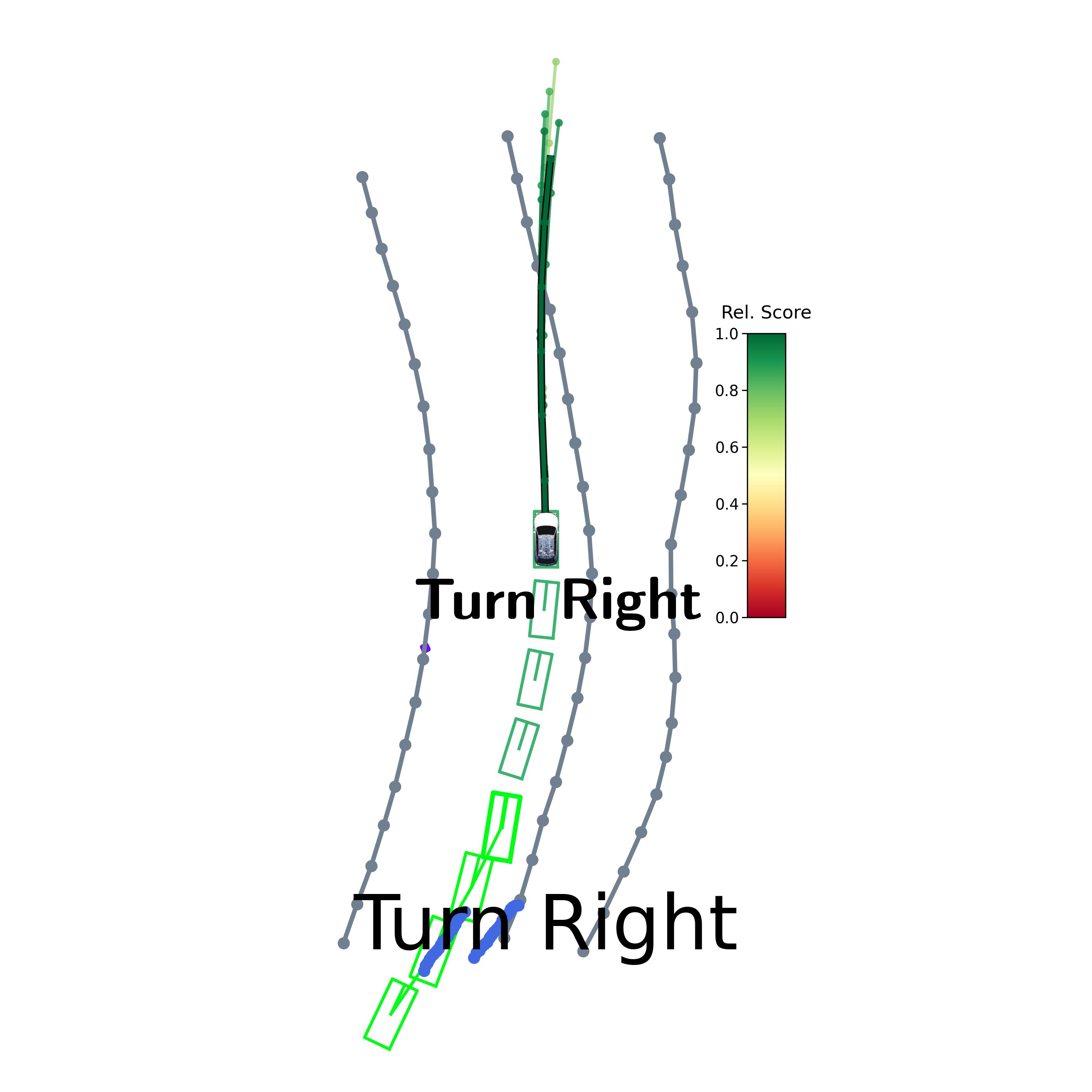}
        \caption{SparseDrive BEV}
        \label{fig:wrong_command_base_bev}
    \end{subfigure}
    \hfill
    \begin{subfigure}[t]{0.23\textwidth}
        \centering
            \includegraphics[
                height=0.97\linewidth,
                trim={800bp 1245bp 580bp 160bp},
                clip
            ]{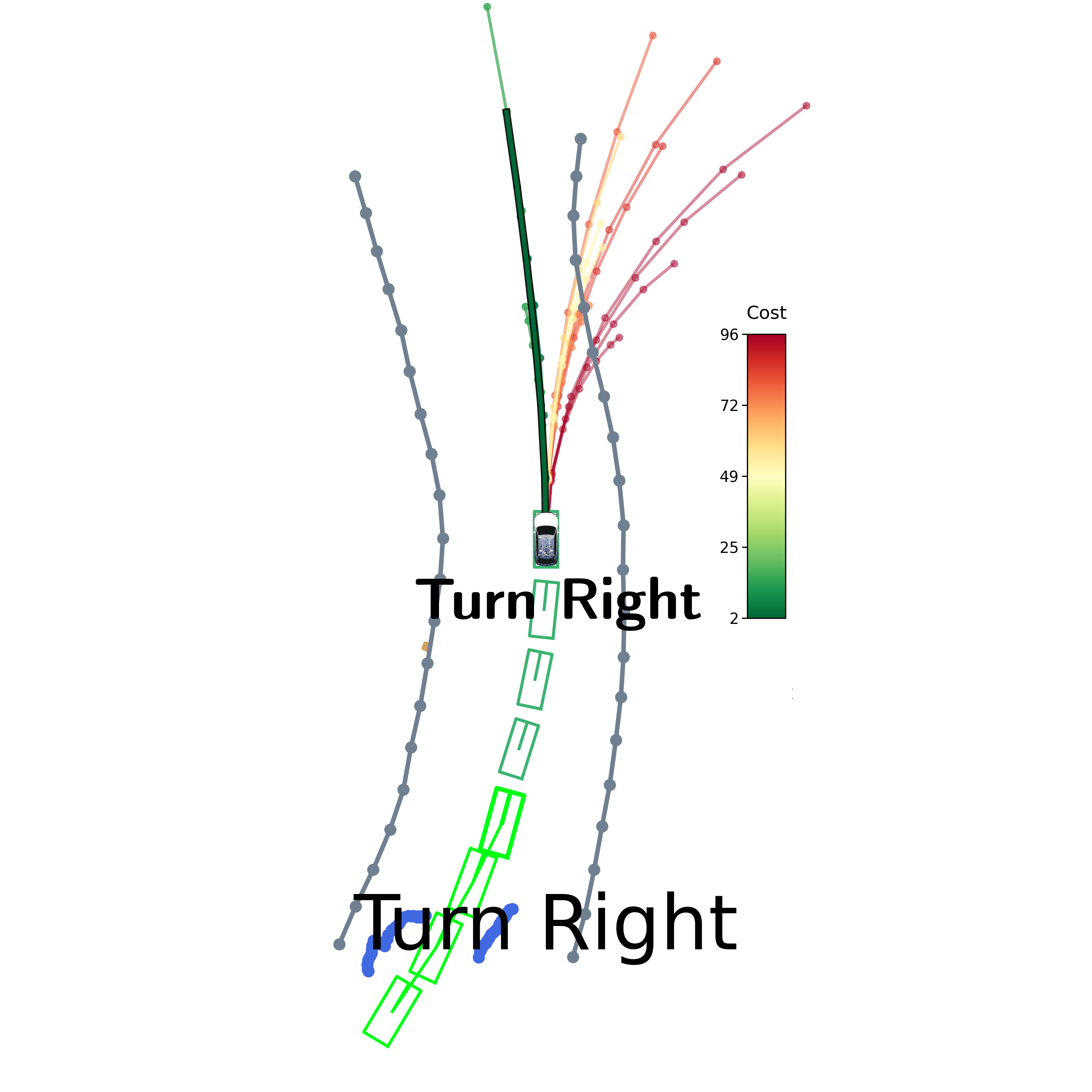}
        \caption{Ours BEV}
        \label{fig:wrong_command_ours_bev}
    \end{subfigure}
    \caption{
    Qualitative example under a temporally premature right-turn command.
    Panel (a) shows the final selected trajectory candidates projected into the camera views, with ours in green and SparseDrive in red.
    SparseDrive follows the premature command into the lane divider, while our planner selects the locally safe left bend.
    }
    \label{fig:wrong_command}
\end{figure*}

Figure~\ref{fig:wrong_command} qualitatively evaluates the behavior of the planners under a temporally misaligned high-level navigation command.
The scene contains an S-bend before a highway merge: the route will soon require a right turn, but the vehicle must first follow a left-hand bend bounded by lane dividers.
Thus, the command is not semantically wrong, but executing the right-turn intent immediately would cut into the divider.
SparseDrive prematurely shifts its selected trajectory toward the upcoming right maneuver and collides with the divider, as seen in Fig.~\ref{fig:wrong_command_base_bev}.
This reflects a limitation of command-conditioned trajectory regression: the generated proposal follows the route intent in geometry, but does not explicitly optimize a local road-compliance cost.

Our planner samples command-conditioned reachable trajectories with the kinematic rollout sampler and evaluates them using the learned query-cost landscape.
Trajectories that conflict with the current lane geometry receive high learned cost, so the selected trajectory follows the low-cost, road-compliant option through the left bend before later committing to the rightward route intent, as illustrated in Fig.~\ref{fig:wrong_command_ours_bev}.
This example shows that query-based cost estimation can use navigation commands while still selecting the safest locally feasible trajectory in the candidate set.


\section{Conclusion}
\label{sec:conc}
This work revisits cost-based planning as an end-to-end driving interface by learning bounded costs over dynamically reachable ego trajectories rather than dense BEV grids.
This shifts supervision toward actions the vehicle can actually execute, 
while retaining a cost representation that can be inspected, optimized, and repaired.
Compact joint scene tokens, safety-aware ranking, and cost-guided cluster mixing support reasoning over coherent future contingencies and produce safe ego plans.
On nuScenes, the approach substantially improves over previous cost-based planners and remains competitive with strong regression-based approaches.
Zero-shot transfer to FZI-AURA demonstrates effectiveness without
fine-tuning, while the cost interface provides a mechanism for inference-time repair using external safety signals.

\section*{Acknowledgements}
The research leading to these results is funded by the German Federal Ministry for Economic Affairs and Energy within the project “NXT GEN AI METHODS – Generative Methoden für Perzeption, Prädiktion und Planung". 
The authors would like to thank the consortium for the successful cooperation. 
The authors acknowledge support by the state of Baden-Württemberg through bwHPC.

%
%
\bibliographystyle{splncs04}
\bibliography{main}

\clearpage
\setcounter{page}{1}
\appendix
\pagebreak
\section*{Supplementary Material for Beyond Waypoint Regression: Query-Based Cost Learning over Reachable Ego Futures for End-to-End Driving}
\addcontentsline{toc}{section}{Appendix}
\label{sec:appendix}

\section{Additional Method Details}
\subsection{Reachable Query Sampling}
\label{app:reachable_query_sampling}

The proposed method uses a clustered set of reachable ego trajectories
$\mathcal{Q}=\{(\tau_i,g_i)\}_{i=1}^{N}$ as the queries of the learned cost landscape.
Here, we describe how these queries are generated before cost evaluation.
The goal is to cover distinct maneuver intentions while keeping every query tied to a physically executable control sequence.
Following sampling-based cost planners such as NMP~\cite{zeng2019end} and ST-P3~\cite{hu2022st}, we use sampled ego trajectories as the support for cost estimation, but organize them as command-conditioned reachable control clusters.
The sampler first defines coarse reference controls for a set of intention clusters, then samples temporally smooth control perturbations around these references.

Given the current ego speed $v_0$ and navigation command $c$, we first define a uniform grid of $N_g$ reference intentions over target final speeds and heading changes.
The target final speeds $\{v_{\mathrm{tar}}^q\}$ span a bounded interval starting from zero up to a command-independent speed limit around the current velocity.
The heading changes $\{\Delta\psi_q\}$ are command-conditioned:
\begin{equation}
    \Delta\psi_q \in
    \begin{cases}
    [-\psi_{\max},0], & c=\mathrm{right},\\
    [0,\psi_{\max}], & c=\mathrm{left},\\
    [-\psi_{\max}/2,\psi_{\max}/2], & c=\mathrm{straight}.
    \end{cases}
    \label{eq:appendix_command_heading_range}
\end{equation}
Each grid point defines a reference intention.
Rather than using it as a geometric anchor, we convert it into constant reference controls over the planning horizon:
\begin{equation}
    a_q^0=\frac{v_{\mathrm{tar}}^q-v_0}{T_{\mathrm{plan}}\Delta t},
    \qquad
    \omega_q^0=\frac{\Delta\psi_q}{T_{\mathrm{plan}}\Delta t},
    \qquad
    \mathbf{u}_{q,t}^{0}=(\omega_q^0,a_q^0).
    \label{eq:appendix_reference_controls}
\end{equation}
Thus, each cluster represents a coarse control intention, such as maintaining speed, slowing down, or turning with different heading changes.

To improve low-speed coverage, the sampler also includes one auxiliary slow-motion cluster.
In total, intra-cluster mixing operates over $\tilde{N}_g=N_g+1$ clusters.
To explore local variations around each intention, we draw $N_p-1$ perturbed control sequences around each cluster control reference and use the auxiliary cluster for the remaining low-speed samples, giving $N=N_g \, N_p$ total queries.
For trajectory $i$ assigned to cluster $g_i$, the sampled control at timestep $t$ is
\begin{equation}
    \mathbf{u}_{i,t}
    =
    \mathbf{u}_{g_i,t}^{0}
    +
    \boldsymbol{\epsilon}_{i,t},
    \qquad
    \mathbf{u}_{i,t}=(\omega_{i,t},a_{i,t}).
    \label{eq:appendix_sampled_controls}
\end{equation}
We use temporally correlated perturbations instead of independent per-step noise, which produces smoother control sequences and avoids unrealistically jittery trajectories:
\begin{equation}
    \boldsymbol{\epsilon}_{i,t}
    =
    \alpha \, \boldsymbol{\epsilon}_{i,t-1}
    +
    \sqrt{1-\alpha^2}\,
    \boldsymbol{\sigma}\odot\mathbf{z}_{i,t},
    \qquad
    \alpha=\exp(-\Delta t/t_c),
    \label{eq:appendix_ar_noise}
\end{equation}
where $\mathbf{z}_{i,t}\sim\mathcal{N}(\mathbf{0},\mathbf{I})$, $\boldsymbol{\sigma}$ sets the yaw-rate and acceleration noise scales, and $t_c$ controls the correlation time.

Finally, each sampled control sequence is rolled out with the kinematic model,
\begin{equation}
    \tau_i=f_{\mathrm{kin}}\!\left(\{\mathbf{u}_{i,t}\}_{t=1}^{T_{\mathrm{plan}}}\right),
    \qquad
    \tau_i=\{\mathbf{x}_{i,t}\}_{t=1}^{T_{\mathrm{plan}}}.
    \label{eq:appendix_query_rollout}
\end{equation}
For clarity, the rollout can be written as a unicycle update for $t=1,\ldots,T_{\mathrm{plan}}$, initialized at the current ego state:
\begin{equation}
    v_{i,t}=v_{i,t-1}+a_{i,t}\Delta t,
    \qquad
    \psi_{i,t}=\psi_{i,t-1}+\omega_{i,t}\Delta t,
    \label{eq:appendix_unicycle_velocity}
\end{equation}
\begin{equation}
    \mathbf{x}_{i,t}
    =
    \mathbf{x}_{i,t-1}
    +
    v_{i,t-1}
    \begin{bmatrix}
    \cos\psi_{i,t-1}\\
    \sin\psi_{i,t-1}
    \end{bmatrix}
    \Delta t
    +
    \frac{1}{2}a_{i,t}
    \begin{bmatrix}
    \cos\psi_{i,t-1}\\
    \sin\psi_{i,t-1}
    \end{bmatrix}
    \Delta t^2 .
    \label{eq:appendix_unicycle_position}
\end{equation}
This corresponding closed-form control integration for nonzero yaw rate is used. The resulting query set is
\begin{equation}
    \mathcal{Q}
    =
    \{(\tau_i,\mathbf{u}_i,g_i)\}_{i=1}^{N},
    \qquad
    N=N_g \, N_p,
    \qquad
    g_i\in\{1,\ldots,\tilde{N}_g\},
    \label{eq:appendix_query_set}
\end{equation}
which is passed to the query-cost planner.

\begin{figure}[h]
    \centering
    \begin{subfigure}[t]{0.98\linewidth}
        \centering
        \includegraphics[width=\linewidth]{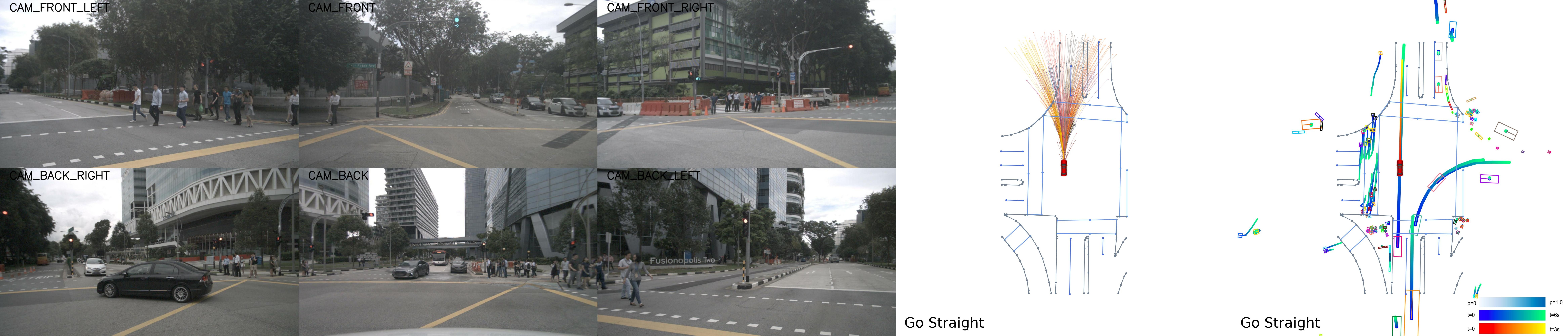}
        \caption{Straight command}
    \end{subfigure}

    \vspace{0.4em}

    \begin{subfigure}[t]{0.98\linewidth}
        \centering
        \includegraphics[width=\linewidth]{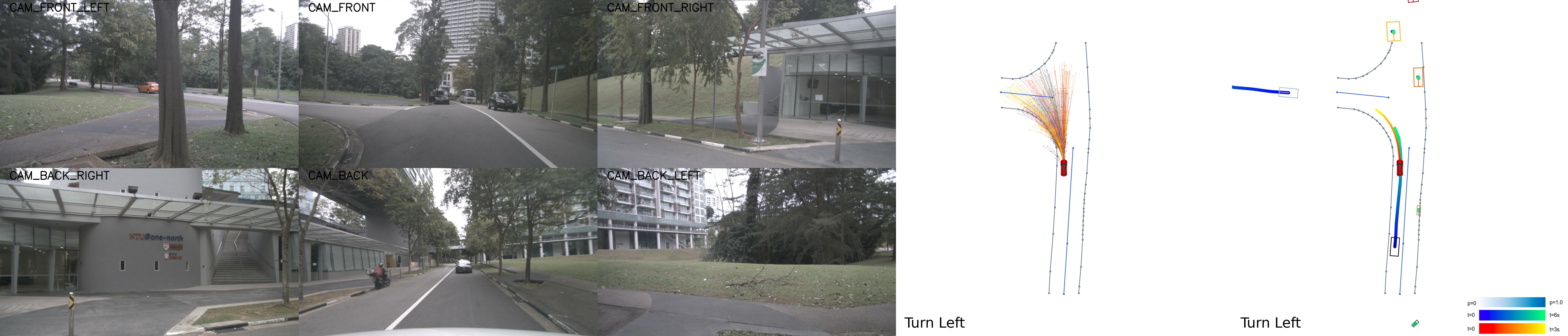}
        \caption{Left command}
    \end{subfigure}

    \vspace{0.4em}

    \begin{subfigure}[t]{0.98\linewidth}
        \centering
        \includegraphics[width=\linewidth]{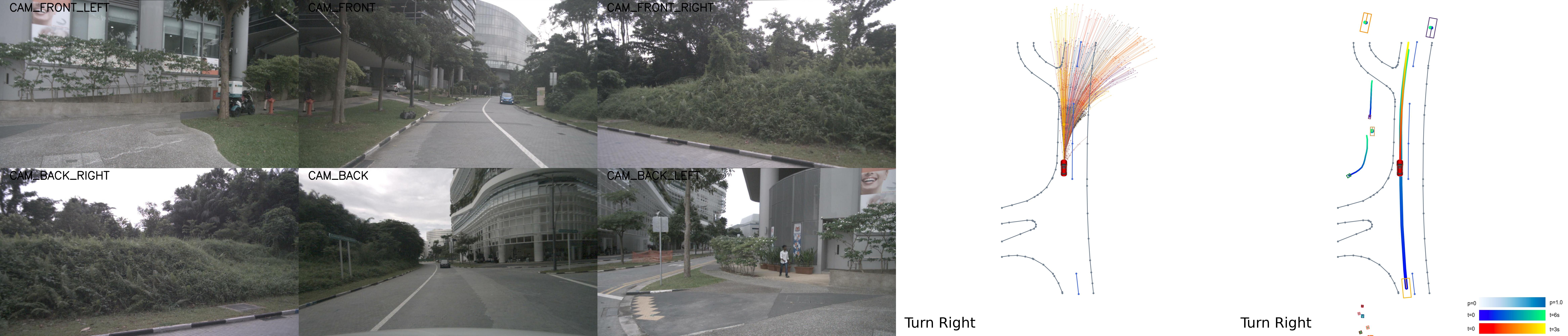}
        \caption{Right command}
    \end{subfigure}
    \caption{
    Command-conditioned reachable query sampling.
    The kinematic sampler generates clustered, dynamically executable ego trajectories that cover the diverse intentions implied by the navigation command while preserving local control variations within each cluster.
    }
    \label{fig:appendix_kinematic_sampler}
\end{figure}

Figure~\ref{fig:appendix_kinematic_sampler} visualizes the resulting command-conditioned query sets for representative straight, left, and right commands from the nuScenes dataset.

\subsection{Cost-Guided Intra-cluster Mixing}
\label{app:cost_guided_cluster_mixing}

Intra-cluster mixing can be written as a weighted average of sampled controls.
Here, we give a more detailed view, which is equivalent to the main paper but makes the MPPI structure explicit.
After contingency aggregation, each sampled trajectory has a per-timestep cost profile $\bar{C}_{i,t}$ and belongs to a mixing cluster $\mathcal{J}_q=\{i:g_i=q\}$, with $q=1,\ldots,\tilde{N}_g$.
For mixing, we define a score profile $R_{i,t}$ from the learned costs.
In the trajectory-level variant, the same score is used at every timestep,
\begin{equation}
    R_{i,t}
    =
    \frac{1}{T_{\mathrm{plan}}}
    \sum_{t'=1}^{T_{\mathrm{plan}}}\bar{C}_{i,t'},
    \qquad \forall t ,
    \label{eq:appendix_mppi_traj_score}
\end{equation}
whereas in the timestep-level variant we use $R_{i,t}=\bar{C}_{i,t}$ directly.
The first option follows the overall quality of each sampled trajectory, while the second recomputes weights at each timestep, so different samples can dominate different parts of the fused control sequence.

Within each cluster, this score profile is converted into the same MPPI-style softmax weights as defined in the main paper, with the cluster-wise minimum score subtracted for numerical stability.
The temperature $\lambda$ controls the sharpness of the update: smaller values concentrate the refinement around the lowest-cost samples, while larger values average over more of the local cost topology.
Rather than averaging trajectory coordinates, our method fuses control perturbations around the reference control of each cluster, then rolls out the fused controls with the ego kinematic model to keep the refined candidate feasible.
\begin{equation}
    \Delta\mathbf{u}_{i,t}
    =
    \mathbf{u}_{i,t}
    -
    \mathbf{u}_{g_i,t}^{0}
    \label{eq:appendix_control_residual}
\end{equation}
denote its deviation from the reference control of its intention cluster.
The refined control sequence for cluster $q$ is then
\begin{equation}
    \tilde{\mathbf{u}}_{q,t}
    =
    \mathbf{u}_{q,t}^{0}
    +
    \sum_{i\in\mathcal{J}_q}
    w_{i,t}^{(q)}
    \Delta\mathbf{u}_{i,t}.
    \label{eq:appendix_residual_control_fusion}
\end{equation}
Because the weights sum to one within each cluster and timestep, this residual form is equivalent to directly averaging the sampled controls with the same MPPI weights.
The residual form simply highlights that the update is performed around the command-conditioned control intention.
Finally, the refined candidate is obtained by rolling out the fused controls,
\begin{equation}
    \tilde{\tau}_q
    =
    f_{\mathrm{kin}}
    \left(
    \{\tilde{\mathbf{u}}_{q,t}\}_{t=1}^{T_{\mathrm{plan}}}
    \right).
    \label{eq:appendix_mppi_rollout}
\end{equation}
This produces one continuous candidate per intention cluster, preserving maneuver diversity while using the learned cost topology to reduce the discretization error of the sampled query set.

\subsection{Cost-Guided Inference and Repair Algorithm}
\label{app:inference_procedure}

    Algorithm~\ref{alg:cost_guided_inference} summarizes how predicted query costs become a selected ego plan. Contingency aggregation combines scene-mode costs, and optional external penalties modify them before fusion. Mean trajectory costs determine within-cluster weights used to mix controls, roll out candidates, and compute cluster-desirability scores for selection. $T=T_{\mathrm{plan}}$, $b=\lfloor T_b/\Delta t\rfloor$, and $E_{i,t}=0$ disables external repair.

    \begin{algorithm}[ht]
    \caption{Cost aggregation, repair, and intra-cluster fusion}
    \label{alg:cost_guided_inference}
    \begin{algorithmic}[1]
    \setlength{\itemsep}{2pt}

    \Require Sampled controls $\mathbf{u}_{i,t}$, cluster IDs $g_i$,
    query costs $C_{i,j,t}$, scene probabilities $\pi_j$,
    optional penalties $E_{i,t}$
    
    \Parameters Horizon $T$, branch index $b$,
    temperature $\lambda>0$, cost bound $C_{\max}$
    
    \Ensure Selected plan $\hat{\tau}$ and candidates
    $\{(\tilde{\tau}_q,\tilde{C}_q)\}$
    
    \For{each query $i$ and timestep $t=1,\ldots,T$}
        \State $\bar{C}_{i,t}\gets
        \begin{cases}
        \max_j C_{i,j,t}, & t\leq b,\\
        \sum_j\pi_j C_{i,j,t}, & t>b
        \end{cases}$
        \State $\bar{C}_{i,t}\gets
        (1-E_{i,t})\bar{C}_{i,t}+E_{i,t}C_{\max}$
    \EndFor
    
    \State $R_i\gets T^{-1}\sum_{t=1}^{T}\bar{C}_{i,t}$
    for every query $i$
    
    \For{each nonempty cluster $q$}
        \State $\mathcal{J}_q\gets\{i:g_i=q\}$, \quad
        $R_q^{\min}\gets\min_{i\in\mathcal{J}_q}R_i$
        \State $w_i^{(q)}\gets
        \dfrac{\exp(-(R_i-R_q^{\min})/\lambda)}
        {\sum_{m\in\mathcal{J}_q}
        \exp(-(R_m-R_q^{\min})/\lambda)}$,
        $i\in\mathcal{J}_q$
        \State $\tilde{\mathbf{u}}_{q,t}\gets
        \sum_{i\in\mathcal{J}_q}w_i^{(q)}\mathbf{u}_{i,t}$
        for all $t$
        \State $\tilde{\tau}_q\gets
        f_{\mathrm{kin}}(\{\tilde{\mathbf{u}}_{q,t}\}_{t=1}^{T})$
        \State $\tilde{C}_q\gets
        \sum_{i\in\mathcal{J}_q}w_i^{(q)}R_i$
    \EndFor
    
    \State $q^\star\gets\arg\min_q\tilde{C}_q$
    \State \Return $\hat{\tau}=\tilde{\tau}_{q^\star}$,
    $\{(\tilde{\tau}_q,\tilde{C}_q)\}$
    \end{algorithmic}
    \end{algorithm}

    The rollout uses the sampler's initialization and kinematic model. We fuse controls rather than trajectory coordinates, and use the cluster-desirability proxy $\tilde{C}_q$ without re-evaluating fused trajectories through the learned cost head. External penalties therefore influence both candidate construction and selection without retraining. The expert trajectory is used only during training.

\subsection{Network Architecture Details.}
\label{app:network_architecture}

We provide the implementation details of the lightweight cost head used to score the reachable ego queries.
Given trajectory embeddings $\mathbf{r}_{i,t}$, ego context $\mathbf{h}_e$, and scene tokens $\{\mathbf{s}_j\}_{j=1}^{N_s}$, the model forms the cost-query tensor by broadcasting
\[
    \mathbf{z}^{\mathrm{cost}}
    \in
    \mathbb{R}^{B\times N\times N_s\times T_{\mathrm{plan}}\times D},
    \qquad
    \mathbf{z}^{\mathrm{cost}}_{i,j,t}
    =
    \mathbf{r}_{i,t}
    +
    \mathbf{h}_e
    +
    \mathbf{s}_j .
\]
The shared cost head is a two-layer MLP applied to every trajectory, scene mode, and timestep:
\[
    \psi_{\mathrm{cost}}(\mathbf{z})
    =
    \mathbf{W}_2\,\mathrm{ReLU}(\mathbf{W}_1\mathbf{z}+\mathbf{b}_1)+\mathbf{b}_2 .
\]
Its scalar output is mapped to $[C_{\min},C_{\max}]$ using the bounded sigmoid cost mapping described in the method; in our experiments, we set $C_{\min}=0$ and $C_{\max}=100$.

\section{Additional Experimental Setup Details}

\subsection{Real World Driving Logs Details.}
\label{app:data_details}

The real-world evaluation set contains 465 driving logs from FZI-AURA collected with the CoCar NextGen research vehicle in Germany across different times of day, lighting conditions, and traffic densities.
Each log has a duration of 20s, resulting in 2.58\,h of raw driving data and 93{,}000 raw samples at 10\, Hz.
The logs cover different locations, times of day, lighting conditions, and traffic densities. The logs include synchronized camera images, ego poses, calibrated sensor transformations, and both 3D bounding-box annotations and semantic labels at 2Hz keyframes.

We match the nuScenes planning protocol and evaluate on 2Hz keyframes.
Across the 465 logs, the dataset contains 725{,}459 annotated 3D boxes.
The labeled keyframes contain an average of 40 boxes, a median of 33, and a range of 1 to 415.
The dominant categories are cars, bicycles, persons, and trucks. 
Frames near the end of a log are excluded from full-horizon planning metrics when the required 3\,s of future trajectory are unavailable.

\subsection{SparseDrive-format preprocessing.}
\label{app:sparsedrive_real_world_preprocessing}

SparseDrive and our method are evaluated with six surround-view cameras, mapped to the nuScenes-style camera slots.
The five forward and side-facing cameras have a native resolution of $1920 \times 1200$, while the rear wide-angle camera has a native resolution of $2592 \times 2048$.
All image inputs are resized and cropped to $704 \times 256$ pixels.

To reduce visual domain shift caused by different camera mounting heights and fields of view, we use a fixed vertical crop offset.
This crop choice aligns the horizon and road layout more closely with the nuScenes image distribution.
Since SparseDrive is trained in the nuScenes top-lidar reference convention, we define a virtual top-lidar frame on the research vehicle to provide the same planner coordinate interface.
This frame follows the nuScenes axis convention, with the $x$ axis pointing right and the $y$ axis pointing forward; the translation from the virtual top-lidar frame to the vehicle base frame is $(0.94,0.00,1.84)$\,m.

High-level navigation commands are generated from the recorded future ego motion using the same virtual frame.
For each evaluated frame, we transform the ego displacement after 3\,s into the virtual top-lidar frame and apply the SparseDrive lateral-threshold rule.
Displacements greater than 2\,m are labeled \texttt{right}, displacements below $-2$\,m are labeled \texttt{left}, and the remaining frames are labeled \texttt{straight}.
This yields a command distribution of 8\% left, 83.5\% straight, and 8.5\% right.
For Alpamayo, we convert the same command labels into text prompts so both methods receive the same high-level intent.

\subsection{Alpamayo Inference Details.}
\label{app:alpamayo_inference}

We evaluate Alpamayo~1.5~\cite{wang2025alpamayo} as a large imitation-oriented vision-language driving reference rather than as a lightweight real-time planner baseline.
Alpamayo is evaluated using only the front-medium camera.
It receives four image history frames and 16 ego-history states sampled at 10\,Hz, corresponding to 1.6\,s of ego-state history.
In contrast, SparseDrive and our method receive four multi-view frames at 2\,Hz.
The same navigation-command heuristic is used for Alpamayo as for the SparseDrive-format methods, with the discrete labels converted into natural-language instructions: ``Turn left'', ``Go straight'', or ``Turn right''.
Alpamayo inference is run on the 10\,Hz raw logs with stride 5, giving the same 2\,Hz evaluation rate as the labeled keyframes.

Predicted trajectories are generated for a 3\,s horizon and exported into the same SparseDrive-style result format used by the other methods.
Alpamayo 1.5 can sample one or more future trajectories, but the released model does not expose a model-defined trajectory-level cost, confidence, likelihood, or score for ranking those samples. 
Therefore, L2 and collision are computed using the first decoded trajectory.
For candidate-diversity evaluation, we collect 16 sampled trajectories per frame.
The samples are generated sequentially due to GPU memory limitations.
For the real-world transfer table in the main paper, latency is measured with a single generated trajectory per frame.

\subsection{Evaluation Metric Details.}
\label{app:metric_details}

We report the standard open-loop L2 and collision metrics at 1s, 2s, and 3s.
Unless otherwise stated, these metrics are computed on the final selected trajectory for each planner.
SparseDrive selects the trajectory with the highest planning score, whereas our method selects the refined candidate with the lowest aggregated query cost.
Following the uniAD~\cite{hu2023planning} protocol, each reported collision value is cumulative over the 0.5s evaluation steps up to the corresponding horizon.
Collision is evaluated by intersecting the predicted ego footprint with future annotated 3D bounding boxes in the ego-aligned evaluation frame.
Map-based road-compliance signals are used outside the reported open-loop L2 and collision metrics.
On nuScenes, boundary violations are used during training to construct the safety-aware ranking margin: a violation is counted when the predicted ego trajectory, evaluated with footprint offsets, crosses drivable-area boundary polylines extracted from lane and road-segment polygons.
On the real-world logs, where the same HD-map protocol is not available, we build an accumulated semantic-lidar road-support map for qualitative visualization and sanity checks.

Additionally, we report candidate-level metrics for diversity and learned cost separation.
For a fixed scene $s$, let $\{\tilde{\tau}_q\}_{q=1}^{K_s}$ denote the refined candidate trajectories, where $\tilde{\tau}_q=\{\tilde{\mathbf{x}}_{q,t}\}_{t=1}^{T_{\mathrm{plan}}}$.

\paragraph{Candidate diversity.}

We measure candidate diversity as the pairwise displacement between candidates produced for the same scene.
We report both an average-displacement version and an endpoint-displacement version:
\[
    \mathrm{Div}_{\mathrm{Avg}}(s)
    =
    \frac{2}{K_s(K_s-1)}
    \sum_{q<q'}
    \frac{1}{T_{\mathrm{plan}}}
    \sum_{t=1}^{T_{\mathrm{plan}}}
    \|\tilde{\mathbf{x}}_{q,t}-\tilde{\mathbf{x}}_{q',t}\|_2 ,
\]
\[
    \mathrm{Div}_{\mathrm{Endpoint}}(s)
    =
    \frac{2}{K_s(K_s-1)}
    \sum_{q<q'}
    \|\tilde{\mathbf{x}}_{q,T_{\mathrm{plan}}}-\tilde{\mathbf{x}}_{q',T_{\mathrm{plan}}}\|_2 .
\]
We average both values over scenes.
$\mathrm{Div}_{\mathrm{Avg}}$ captures the average spread of the candidates over the full planning horizon, while $\mathrm{Div}_{\mathrm{Endpoint}}$ emphasizes endpoint separation and more directly reflects diversity between maneuver intentions.

\paragraph{Cost separation.}

For cost-based variants, we evaluate whether the learned costs separate unsafe candidates from safe candidates.
A candidate is unsafe if it collides with another actor or leaves the drivable area at any valid timestep.
Let $\mathcal{U}_s$ and $\mathcal{R}_s$ denote the unsafe and safe refined-candidate index sets in scene $s$, respectively, and let $\tilde{C}_q$ denote the final cluster cost of refined candidate $\tilde{\tau}_q$.
Following the standard probabilistic interpretation of ROC-AUC~\cite{fawcett2006introduction}, cost AUC measures how often an unsafe candidate is ranked above a safe candidate.
Since higher costs correspond to less desirable trajectories, it estimates $\Pr(\tilde{C}_u>\tilde{C}_r)+\frac{1}{2}\Pr(\tilde{C}_u=\tilde{C}_r)$, where $u\sim\mathcal{U}_s$ and $r\sim\mathcal{R}_s$.
For scenes containing at least one safe and one unsafe candidate, we compute this empirical AUC using the following ranking form:
\[
    \mathrm{AUC}_s
    =
    \frac{1}{|\mathcal{U}_s||\mathcal{R}_s|}
    \sum_{u\in\mathcal{U}_s}
    \sum_{r\in\mathcal{R}_s}
    \left[
    \indicator_{\{\tilde{C}_u > \tilde{C}_r\}}
    +
    \frac{1}{2}\, \indicator_{\{\tilde{C}_u=\tilde{C}_r\}}
    \right].
\]
The equality term follows the standard half-credit convention for ties in AUC; exact equality is typically rare for continuous learned costs, but the term makes the metric well-defined for bounded or numerically equal scores.
Thus, an AUC of $0.5$ indicates random ordering, while $1.0$ indicates perfect separation.
We also report the cost gap,
\[
    \mathrm{Gap}_s
    =
    \frac{1}{|\mathcal{U}_s|}
    \sum_{u\in\mathcal{U}_s} \tilde{C}_u
    -
    \frac{1}{|\mathcal{R}_s|}
    \sum_{r\in\mathcal{R}_s} \tilde{C}_r .
\]
Larger positive gaps indicate that unsafe candidates receive substantially higher costs than safe candidates on average.
The reported AUC and gap are averaged over scenes where both safe and unsafe candidates exist.

\section{Additional Experiments}

\subsection{Branching Time.}
\label{app:branching_time}
Table~\ref{tab:appendix_branching_time} evaluates the branching time that separates worst-case from probability-weighted scene-cost aggregation.
At $T_b=0$, probability-weighted aggregation is used throughout, while at $T_b=3$\,s, worst-case aggregation covers the full planning horizon.
Pure expectation can underweight low-probability, high-cost futures, whereas a long worst-case interval can favor conservative responses even when the corresponding scene mode is unlikely.

The default $T_b=1$\,s achieves both the lowest average L2 error ($0.71$\,m) and the lowest average collision rate ($0.07\%$) among the ablated branching times.
Extending the worst-case interval to two or three seconds degrades both metrics, so more conservative cost aggregation does not automatically yield fewer collisions.
These results support a short worst-case prefix for near-term decisions, followed by probability-weighted costs that distinguish the likelihood of later outcomes.
\begin{table}[ht]
    \centering
    \caption{Ablation of Branching-time on nuScenes.}
    \label{tab:appendix_branching_time}
    \resizebox{0.7\linewidth}{!}{%
    \setlength{\tabcolsep}{3.2pt}%
    \begin{tabular}{@{}lccc>{\columncolor[gray]{0.88}}cccc>{\columncolor[gray]{0.88}}c@{}}
        \toprule
        \multirow{2}{*}{$T_b$ (s)} &
        \multicolumn{4}{c}{L2 (m) $\downarrow$} &
        \multicolumn{4}{c}{Collision (\%) $\downarrow$} \\
        \cmidrule(lr){2-5}\cmidrule(lr){6-9}
        & 1s & 2s & 3s & Avg. & 1s & 2s & 3s & Avg. \\
        \midrule
        0 & 0.33 & 0.80 & 1.50 & 0.88 & 0.02 & 0.14 & 0.35 & 0.17 \\
        1 & 0.28 & 0.65 & 1.21 & \textbf{0.71} & 0.00 & 0.02 & 0.18 & \textbf{0.07} \\
        2 & 0.30 & 0.69 & 1.28 & 0.76 & 0.02 & 0.07 & 0.29 & 0.13 \\
        3 & 0.32 & 0.78 & 1.47 & 0.86 & 0.03 & 0.16 & 0.37 & 0.19 \\
        \bottomrule
    \end{tabular}
    }
\end{table}

\subsection{Plan Selection and Rescoring.}
\label{app:selection_rescoring}

Table~\ref{tab:appendix_selection_rescoring} compares global lowest-cost selection, best-per-cluster selection, MPPI mixing, and MPPI mixing with learned rescoring of the fused trajectories.
Global selection chooses the lowest-cost sample without preserving a candidate for each intention, while best-per-cluster selection maintains a representative sample from each cluster.
MPPI instead uses the learned safety-aware costs to mix controls within each intention cluster and rolls out the fused controls with the kinematic model, allowing refinement beyond individual samples without averaging distinct maneuver families.

Compared with global selection, MPPI reduces average L2 from $0.77$ to $0.71$\,m and collision from $0.13\%$ to $0.07\%$.
Best-per-cluster selection achieves lower L2 ($0.68$\,m), but its higher collision rate ($0.11\%$) highlights the separation between matching the expert trajectory and avoiding collisions.
The mixing results therefore support using the learned cost topology for safety-aware refinement rather than only selecting fixed samples.

The rescoring experiment tests whether evaluating the fused trajectory with the learned cost head improves selection over the cluster-desirability proxy.
The proxy combines sample costs using the fusion weights, whereas rescoring directly evaluates the resulting trajectory.
Rescoring increases average L2 from $0.71$ to $0.73$\,m and average collision from $0.07\%$ to $0.15\%$ relative to the default.
Empirical analysis shows that the cluster-desirability proxy is strongly correlated with the learned cost evaluated directly on the mixed trajectory, with Pearson and Spearman correlation coefficients of $0.921$ and $0.923$, respectively.
\begin{table}[ht]
    \centering
    \caption{Ablation of Plan selection and rescoring on nuScenes.}
    \label{tab:appendix_selection_rescoring}
    \resizebox{0.7\linewidth}{!}{%
    \setlength{\tabcolsep}{3.2pt}%
    \begin{tabular}{@{}lccc>{\columncolor[gray]{0.88}}cccc>{\columncolor[gray]{0.88}}c@{}}
        \toprule
        \multirow{2}{*}{Variant} &
        \multicolumn{4}{c}{L2 (m) $\downarrow$} &
        \multicolumn{4}{c}{Collision (\%) $\downarrow$} \\
        \cmidrule(lr){2-5}\cmidrule(lr){6-9}
        & 1s & 2s & 3s & Avg. & 1s & 2s & 3s & Avg. \\
        \midrule
        Global selection & 0.29 & 0.70 & 1.30 & 0.77 & 0.02 & 0.07 & 0.31 & 0.13 \\
        Best per cluster & 0.26 & 0.63 & 1.14 & \textbf{0.68} & 0.01 & 0.05 & 0.28 & 0.11 \\
        MPPI & 0.28 & 0.65 & 1.21 & 0.71 & 0.00 & 0.02 & 0.18 & \textbf{0.07} \\
        MPPI + rescoring & 0.28 & 0.67 & 1.22 & 0.73 & 0.01 & 0.12 & 0.31 & 0.15 \\
        \bottomrule
    \end{tabular}
    }
\end{table}

\subsection{Planning Loss Components.}
Table~\ref{tab:appendix_loss_components} ablates the planner-specific losses while keeping the same reachable-query architecture.
The safety margin corresponds to $\mathcal{L}_{\mathrm{rank}}$, while candidate classification and candidate regression correspond to $\mathcal{L}_{\mathrm{cls}}$ and $\mathcal{L}_{\mathrm{reg}}$ of the training objective $\mathcal{L}_{\mathrm{plan}}$. The safety-margin ranking loss alone already learns a meaningful safe-versus-unsafe ordering, as reflected by a cost AUC of 0.81 and a cost gap of 36.17, but the final plan remains limited by weak cluster-level supervision.
Adding candidate classification improves displacement error by encouraging the planner to select a more expert-like intention cluster, although this alone does not improve collision.
Adding candidate regression reduces collision relative to the safety-margin-only variant, but without the classification loss, it does not reliably select the best cluster and therefore hurts displacement.
Using all three losses gives the best overall behavior: the ranking loss shapes the cost topology, classification improves inter-cluster selection, and regression refines the selected candidate geometry.

\begin{table}[h]
    \centering
    \caption{Ablation of planning loss components on nuScenes.}
    \label{tab:appendix_loss_components}
    \resizebox{0.95\linewidth}{!}{%
    \setlength{\tabcolsep}{3.2pt}%
    \begin{tabular}{@{}lccc>{\columncolor[gray]{0.88}}cccc>{\columncolor[gray]{0.88}}ccc@{}}
        \toprule
        \multirow{2}{*}{Variant} &
        \multicolumn{4}{c}{L2 (m) $\downarrow$} &
        \multicolumn{4}{c}{Collision (\%) $\downarrow$} &
        \multirow{2}{*}{Cost AUC $\uparrow$} &
        \multirow{2}{*}{Cost Gap $\uparrow$} \\
        \cmidrule(lr){2-5}\cmidrule(lr){6-9}
        & 1s & 2s & 3s & Avg. & 1s & 2s & 3s & Avg. & & \\
        \midrule
        Safety margin                         & 0.35 & 0.89 & 1.72 & 0.99 & 0.00 & 0.151 & 0.41 & 0.19 & 0.81 & 36.17 \\
        + candidate classification            & 0.34 & 0.82 & 1.54 & 0.90 & 0.04 & 0.19 & 0.44 & 0.22 & 0.81 & 38.64 \\
        + candidate regression                & 0.38 & 1.00 & 1.91 & 1.10 & 0.04 & 0.11 & 0.31 & 0.15 & 0.82 & 35.76 \\
        + classification + regression         & 0.28 & 0.65 & 1.21 & 0.71 & 0.00 & 0.02 & 0.18 & 0.07 & 0.86 & 45.70  \\
        \bottomrule
    \end{tabular}
    }
\end{table}

\subsection{Hard-Negative Aggregation.}

Table~\ref{tab:appendix_hard_negative} studies two choices in the safety-ranking loss: the collision signal used to define the margin, and how hard negatives are aggregated.
The max variant backpropagates only through the single largest ranking violation in each scene, while the Top-$k$ variant averages the largest $k_h$ violations.
Thus, max aggregation focuses on the hardest negative but can produce sparse gradients, whereas Top-$k$ spreads supervision over multiple unsafe or near-unsafe candidates.

For ground-truth collision supervision, Top-$k$ aggregation consistently improves the learned cost topology and final planning, reducing L2 from 0.76 to 0.69 and increasing the cost gap from 41.67 to 46.66.
A similar trend appears for contingency collision risk, where Top-$k$ aggregation reduces the average collision rate from 0.13\% to 0.07\% and improves both cost AUC and cost gap.
This suggests that when the collision signal is reliable or contingency-aware, spreading gradients across several hard negatives yields a smoother and more useful cost topology than optimizing only the worst single violation.

The top-1 predicted collision signal behaves differently: max aggregation yields a lower collision rate than Top-$k$, likely because a single predicted future provides a narrower, noisier safety target.
In that setting, averaging many hard negatives can overemphasize candidates that are unsafe only under the most likely predicted future, rather than under the broader multimodal risk.
Overall, the best result is obtained by combining contingency collision risk with Top-$k$ aggregation, supporting our default choice of supervising the ranking loss with multiple informative negatives under multimodal scene risk.

\begin{table}[h]
    \centering
    \caption{Ablation of hard-negative aggregation for safety-margin training on nuScenes.}
    \label{tab:appendix_hard_negative}
    \resizebox{0.95\linewidth}{!}{%
    \setlength{\tabcolsep}{3.2pt}%
    \begin{tabular}{@{}llccc>{\columncolor[gray]{0.88}}cccc>{\columncolor[gray]{0.88}}ccc@{}}
        \toprule
        \multirow{2}{*}{Collision signal} &
        \multirow{2}{*}{Negative agg.} &
        \multicolumn{4}{c}{L2 (m) $\downarrow$} &
        \multicolumn{4}{c}{Collision (\%) $\downarrow$} &
        \multirow{2}{*}{Cost AUC $\uparrow$} &
        \multirow{2}{*}{Cost Gap $\uparrow$} \\
        \cmidrule(lr){3-6}\cmidrule(lr){7-10}
        & & 1s & 2s & 3s & Avg. & 1s & 2s & 3s & Avg. & & \\
        \midrule
        Ground-truth collision      & Max   & 0.29 & 0.69 & 1.30 & 0.76 & 0.01 & 0.06 & 0.27 & 0.11  & 0.85 & 41.67 \\
        Ground-truth collision      & Top-$k$ & 0.27 & 0.63 & 1.16 & 0.69 & 0.00 & 0.06 & 0.25 & 0.10 & 0.86 & 46.66 \\
        Top-1 predicted collision   & Max   & 0.27 & 0.62 & 1.16 & 0.68 & 0.00 & 0.04 & 0.19 & 0.08 & 0.88 & 42.56 \\
        Top-1 predicted collision   & Top-$k$ & 0.27 & 0.61 & 1.12 & 0.67 & 0.04 & 0.12 & 0.23 & 0.13 & 0.87 & 45.71 \\
        Contingency collision risk  & Max   & 0.28 & 0.66 & 1.23 & 0.72 & 0.01 & 0.10 & 0.28 & 0.13 & 0.84 & 41.22 \\
        Contingency collision risk & Top-$k$ & 0.28 & 0.65 & 1.21 & 0.71 & 0.00 & 0.02 & 0.18 & 0.07 & 0.86 & 45.70 \\
        \bottomrule
    \end{tabular}
    }
\end{table}

\subsection{Cost Profile for Cluster Mixing.}

Table~\ref{tab:appendix_mppi_cost_profile} evaluates which learned cost profile is used to compute the MPPI-style cluster weights.
The mean trajectory cost first averages the cost over the horizon and uses a single score for each sampled trajectory, giving the refinement a trajectory-level preference.
The per-timestep cost instead computes weights independently at each future step, which is more local but can combine controls from different samples at different timesteps.

The two variants obtain the same cost AUC, indicating a similar safe-versus-unsafe ordering.
The difference appears in how this topology is used for refinement: the mean trajectory cost reduces the average L2 from 0.77 to 0.71 and the average collision from 0.14\% to 0.07\%.
This suggests that intra-cluster mixing benefits from a temporally consistent trajectory-level preference.
Per-timestep weighting is more flexible, but in a finite sampled cluster, it can assemble a control sequence from samples that are locally low-cost at different times but do not form a coherent maneuver.
We therefore use the mean trajectory cost as the default profile for cost-guided intra-cluster mixing.

\begin{table}[h]
    \centering
    \caption{Ablation of cost profiles for cost-guided cluster mixing on nuScenes.}
    \label{tab:appendix_mppi_cost_profile}
    \resizebox{0.95\linewidth}{!}{%
    \setlength{\tabcolsep}{3.2pt}%
    \begin{tabular}{@{}lccc>{\columncolor[gray]{0.88}}cccc>{\columncolor[gray]{0.88}}ccc@{}}
        \toprule
        \multirow{2}{*}{Mixing score} &
        \multicolumn{4}{c}{L2 (m) $\downarrow$} &
        \multicolumn{4}{c}{Collision (\%) $\downarrow$} &
        \multirow{2}{*}{Cost AUC $\uparrow$} &
        \multirow{2}{*}{Cost Gap $\uparrow$} \\
        \cmidrule(lr){2-5}\cmidrule(lr){6-9}
        & 1s & 2s & 3s & Avg. & 1s & 2s & 3s & Avg. & & \\
        \midrule
        Mean trajectory cost   & 0.28 & 0.65 & 1.21 & 0.71 & 0.00 & 0.02 & 0.18 & 0.07 & 0.86 & 45.70 \\
        Per-timestep cost      & 0.30 & 0.71 & 1.32 & 0.77 & 0.02 & 0.10 & 0.30 & 0.14 & 0.86 & 42.99 \\
        \bottomrule
    \end{tabular}
    }
\end{table}

\subsection{Scene-Cost Aggregation.}
Table~\ref{tab:appendix_contingency_aggregation} ablates how the scene-mode costs are aggregated before cluster mixing.
Top-1 aggregation uses only the most likely scene mode, worst-case aggregation takes the maximum over scene modes, mean aggregation treats all modes uniformly, and probability-weighted aggregation follows the predicted scene probabilities.
The contingency strategy combines these views temporally: it uses worst-case costs before the branching time and probability-weighted costs afterward.

The aggregation strategy has a clear effect on both planning accuracy and safety.
Top-1 and probability-weighted aggregation commit strongly to likely scene modes and can therefore underemphasize low-probability but safety-critical futures.
Worst-case aggregation avoids this optimism, but applying the maximum over all scene modes for the full horizon can be overly conservative and does not improve collision in this setting.
Uniform mean aggregation provides a moderate compromise, reducing collision relative to Top-1 and worst-case aggregation, but it ignores the different roles of early and late planning timesteps.

The contingency aggregation performs best across all metrics.
By using worst-case costs before the branching time, it keeps early ego actions safe under multiple plausible futures.
After the branching time, probability-weighted aggregation allows the planner to account for the likelihood of each scene mode.
This temporal structure improves average L2 from 0.85--0.88 to 0.71 and reduces collision from 0.13--0.19\% to 0.07\%, while producing the strongest cost separation.

\begin{table}[h]
    \centering
    \caption{Ablation of scene-cost aggregation on nuScenes.}
    \label{tab:appendix_contingency_aggregation}
    \resizebox{0.95\linewidth}{!}{%
    \setlength{\tabcolsep}{3.2pt}%
    \begin{tabular}{@{}lccc>{\columncolor[gray]{0.88}}cccc>{\columncolor[gray]{0.88}}ccc@{}}
        \toprule
        \multirow{2}{*}{Aggregation} &
        \multicolumn{4}{c}{L2 (m) $\downarrow$} &
        \multicolumn{4}{c}{Collision (\%) $\downarrow$} &
        \multirow{2}{*}{Cost AUC $\uparrow$} &
        \multirow{2}{*}{Cost Gap $\uparrow$} \\
        \cmidrule(lr){2-5}\cmidrule(lr){6-9}
        & 1s & 2s & 3s & Avg. & 1s & 2s & 3s & Avg. & & \\
        \midrule
        Top-1                  & 0.33 & 0.79 & 1.50 & 0.87 & 0.03 & 0.17 & 0.37 & 0.19 & 0.81 & 33.99 \\
        Worst Case             & 0.32 & 0.78 & 1.47 & 0.86 & 0.03 & 0.16 & 0.37 & 0.19 & 0.81 & 37.81 \\
        Mean                   & 0.32 & 0.77 & 1.45 & 0.85 & 0.00 & 0.09 & 0.30 & 0.13 & 0.82 & 38.97 \\
        Probability-weighted   & 0.33 & 0.80 & 1.50 & 0.88 & 0.02 & 0.14 & 0.35 & 0.17 & 0.81 & 37.97 \\
        \textbf{Contingency}   & 0.28 & 0.65 & 1.21 & 0.71 & 0.00 & 0.02 & 0.18 & 0.07 & 0.86 & 45.70 \\
        \bottomrule
    \end{tabular}
    }
\end{table}
\section{Additional Qualitative Results}

\subsection{nuScenes Examples.}

\begin{figure}[ht]
    \centering
    \begin{subfigure}[t]{0.94\textwidth}
        \centering
        \includegraphics[width=\linewidth]{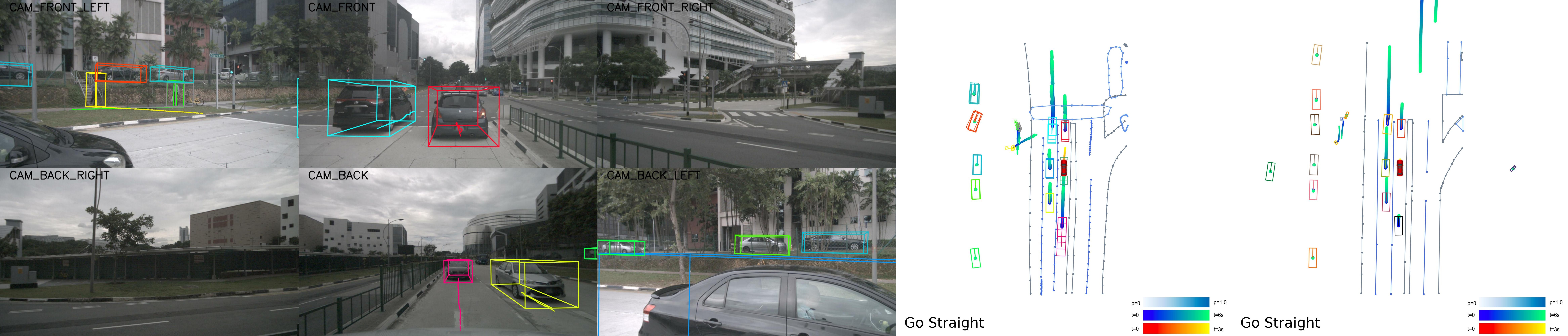}
        \caption{Following a leading vehicle}
    \end{subfigure}

    \vspace{0.5em}

    \begin{subfigure}[t]{0.94\textwidth}
        \centering
        \includegraphics[width=\linewidth]{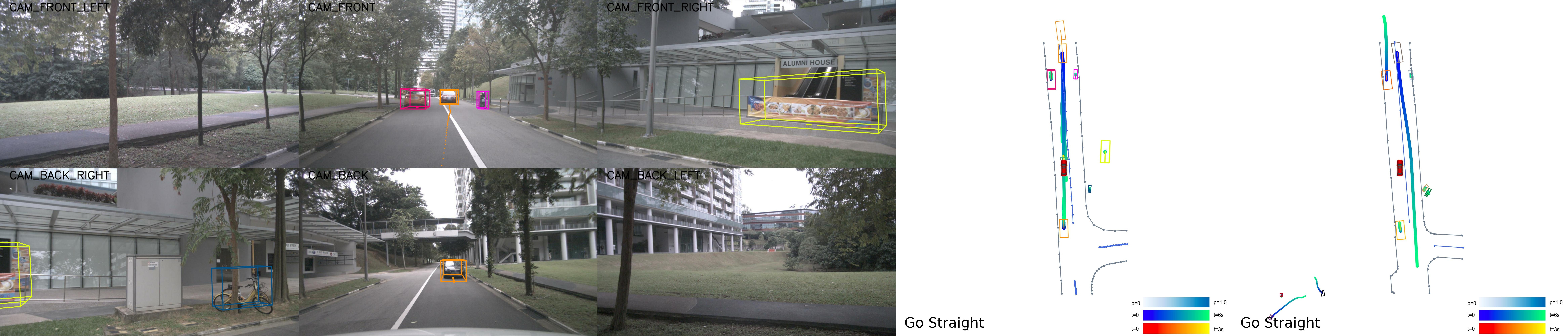}
        \caption{Yielding to oncoming traffic}
    \end{subfigure}

    \vspace{0.5em}

    \begin{subfigure}[t]{0.94\textwidth}
        \centering
        \includegraphics[width=\linewidth]{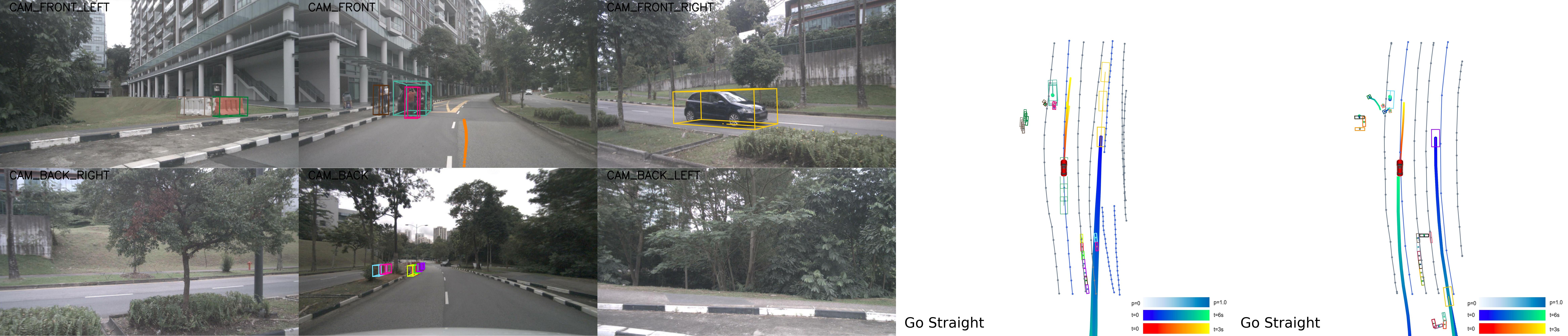}
        \caption{Lane change Maneuver}
    \end{subfigure}

    \vspace{0.5em}

    \begin{subfigure}[t]{0.94\textwidth}
        \centering
        \includegraphics[width=\linewidth]{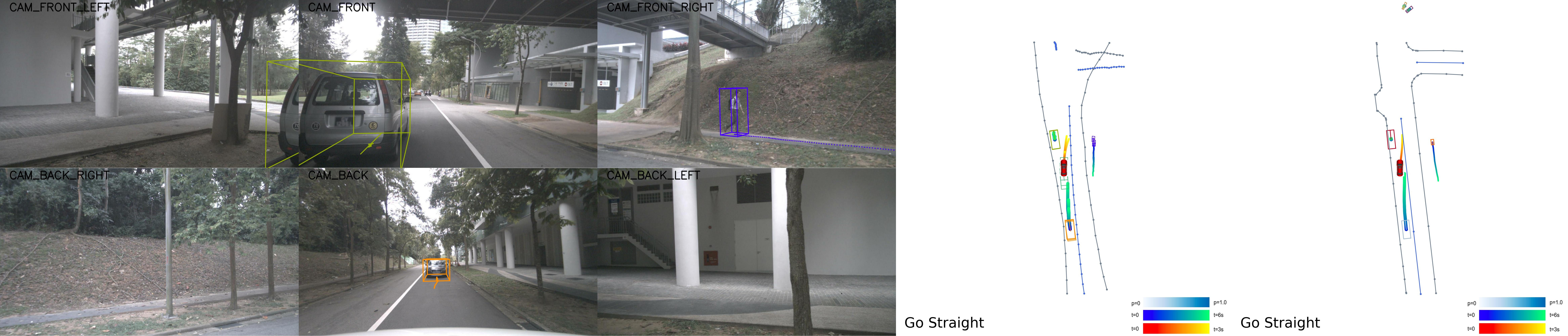}
        \caption{Overtaking Maneuver}
    \end{subfigure}
    \caption{
    Qualitative nuScenes examples.
    The camera views and BEV visualizations show representative interactive scenes in which the learned cost topology guides selection among reachable ego candidates.
    }
    \label{fig:appendix_nuscenes_examples}
\end{figure}

Figure~\ref{fig:appendix_nuscenes_examples} shows representative scenes covering car following, yielding, lane-change, and overtaking behavior from nuScenes.
These examples complement the quantitative results by illustrating how the reachable-query planner evaluates a diverse set of ego futures in interactive traffic scenes and selects low-cost candidates that remain consistent with the local road geometry and surrounding agents.

\subsection{Zero-shot Real-World Examples}
\begin{figure}[ht]
    \centering
    \begin{subfigure}[t]{0.94\textwidth}
        \centering
        \includegraphics[width=\linewidth]{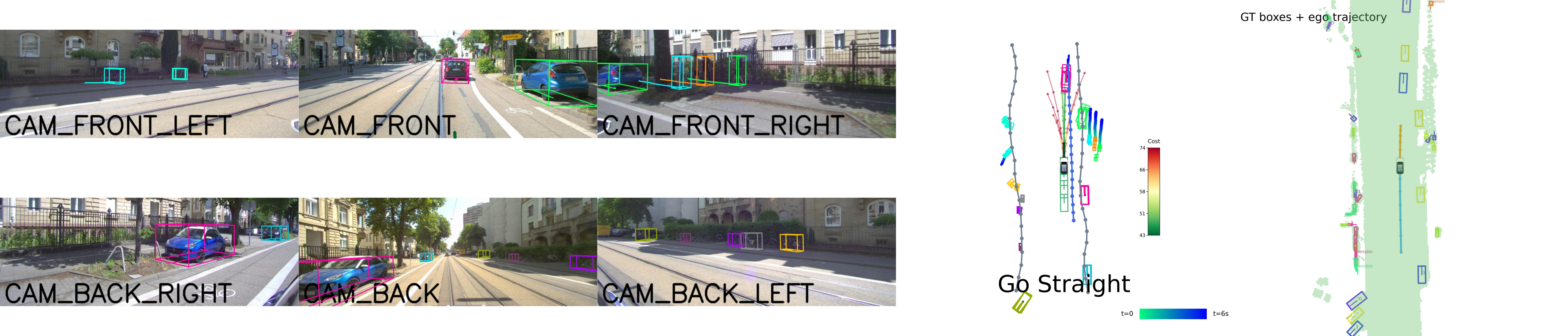}
        \caption{Reacting to a stopping lead vehicle}
    \end{subfigure}

    \vspace{0.5em}

    \begin{subfigure}[t]{0.94\textwidth}
        \centering
        \includegraphics[width=\linewidth]{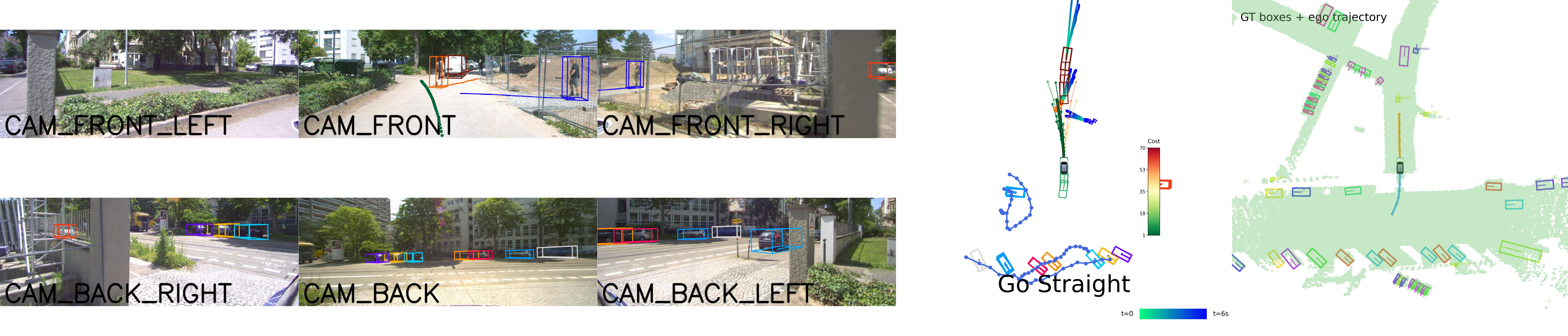}
        \caption{Reacting to a pedestrian}
    \end{subfigure}

    \vspace{0.5em}

    \begin{subfigure}[t]{0.94\textwidth}
        \centering
        \includegraphics[width=\linewidth]{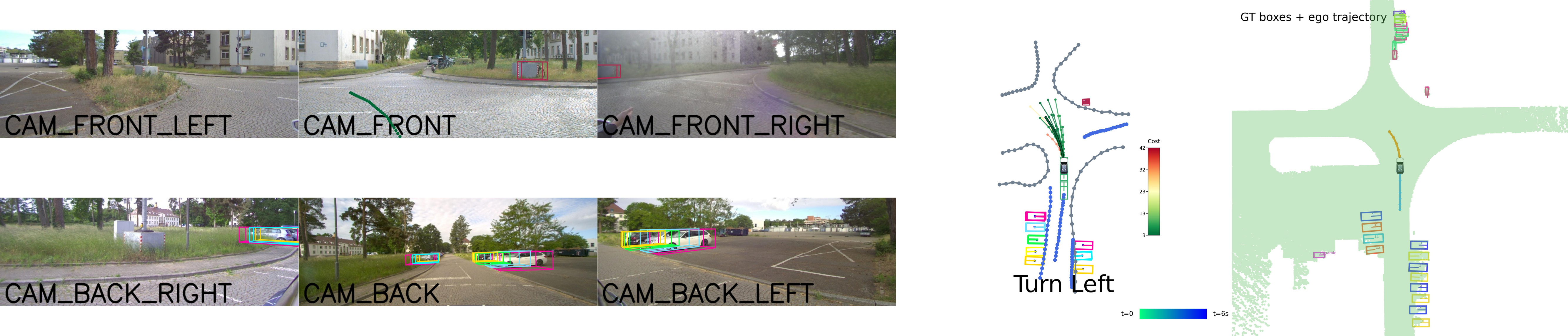}
        \caption{Turning left at an intersection}
    \end{subfigure}

        \vspace{0.5em}

    \begin{subfigure}[t]{0.94\textwidth}
        \centering
        \includegraphics[width=\linewidth]{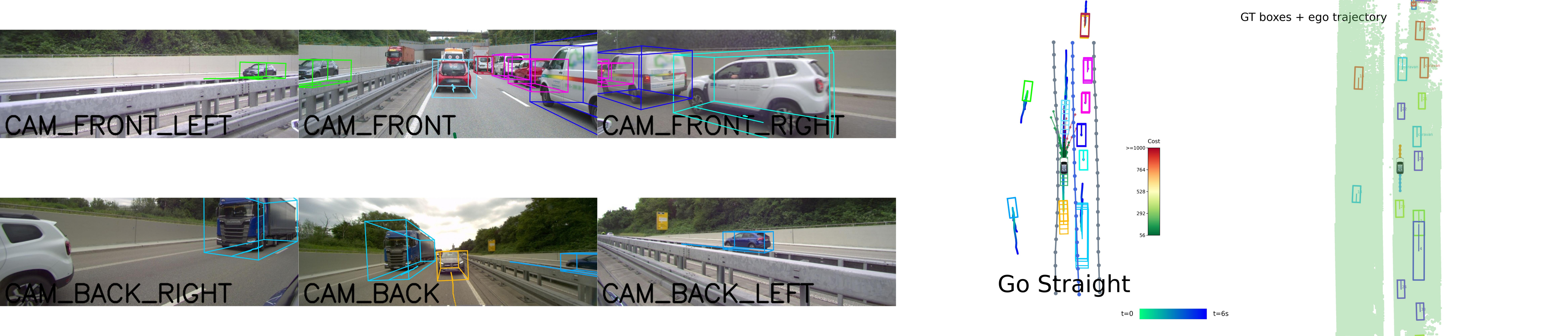}
        \caption{Stop-and-go traffic on a highway}
    \end{subfigure}
    \caption{
    Zero-shot qualitative examples on real-world driving logs. Each row shows surround-view camera predictions, the BEV cost visualization with the selected trajectory, and ground-truth boxes with ego trajectory and semantic map support.
    }
    \label{fig:appendix_ourdata_examples}

\end{figure}

Figure~\ref{fig:appendix_ourdata_examples} shows representative zero-shot transfer examples on our real-world driving logs, covering stop-and-go traffic, interaction with vulnerable road users, intersection turning, and highway following. These examples complement the quantitative real-world evaluation by illustrating that the learned cost topology remains meaningful under a different sensor setup, camera domain, and road layout. Each row shows the surround camera inputs, the BEV cost visualization over top-k reachable ego candidates, and the ground-truth ego trajectory together with semantic map support. In the middle BEV panel, the selected final trajectory candidate is highlighted in dark green with a black outline.


\end{document}